\documentclass[shortpaper]{clv3}

\usepackage{xcolor}
\definecolor{darkblue}{rgb}{0, 0, 0.5}

\usepackage{graphicx}
\usepackage{booktabs}
\usepackage{algorithm}
\usepackage{algpseudocode}
\usepackage{framed}

\usepackage{mathtools}
\usepackage{amsmath}
\usepackage{amssymb}
\DeclareMathOperator*{\argmax}{arg\,max}

\algblock{Input}{EndInput}
\algnotext{EndInput}
\algblock{Output}{EndOutput}
\algnotext{EndOutput}

\definecolor{algoblue}{HTML}{2E4C8F}
\definecolor{algored}{HTML}{CC4844}

\usepackage{hyperref}
\hypersetup{colorlinks=true,citecolor=darkblue,linkcolor=darkblue, urlcolor=darkblue}

\usepackage{todonotes}

\definecolor{major-change}{HTML}{000000} 
\definecolor{major-change-v3}{HTML}{000000} 
\definecolor{major-change-v4}{HTML}{000000} 

\newcommand{\sout}[1]{\unskip} 
\usepackage{soul} 
\renewcommand{\st}[1]{\unskip}

\begin{document}
\issue{1}{1}{2024}


\runningtitle{A Principled Framework for Evaluating on Typologically Diverse Languages}
\runningauthor{Ploeger et al.}


\title{A Principled Framework for Evaluating on Typologically Diverse Languages}

\author{Esther Ploeger}
\affil{Aalborg University \\
Department of Computer Science \\
 \texttt{espl@cs.aau.dk}
}

\author{Wessel Poelman}
\affil{KU Leuven \\
Department of Computer Science \\
 \texttt{wessel.poelman@kuleuven.be}
}

\author{Andreas Holck Høeg-Petersen}
\affil{Aalborg University \\
Department of Computer Science \\
 \texttt{ahhp@cs.aau.dk}
}

\author{Anders Schlichtkrull}
\affil{Aalborg University \\
Department of Computer Science \\
 \texttt{andsch@cs.aau.dk}
}

\author{Miryam de Lhoneux}
\affil{KU Leuven \\
Department of Computer Science \\
 \texttt{miryam.delhoneux@kuleuven.be}
}

\author{Johannes Bjerva}
\affil{Aalborg University \\
Department of Computer Science \\
 \texttt{jbjerva@cs.aau.dk}
}

\maketitle

\begin{abstract}
Beyond individual languages, multilingual natural language processing (NLP) research increasingly aims to develop models that perform well across languages generally.
However, evaluating these systems on all the world's languages is practically infeasible.
To attain generalizability, representative language sampling is essential. 
Previous work argues that generalizable multilingual evaluation sets should contain languages with diverse typological properties.
However, `typologically diverse' language samples have been found to vary considerably in this regard, and popular sampling methods are flawed and inconsistent.
We present a language sampling framework for selecting highly typologically diverse languages given a sampling frame, informed by language typology.
We compare sampling methods with a range of metrics and find that our systematic  methods consistently retrieve more typologically diverse language selections than previous methods in NLP.
Moreover, we provide evidence that this affects generalizability in multilingual model evaluation, emphasizing the importance of diverse language sampling in NLP evaluation.
\end{abstract}

{\let\thefootnote\relax\footnotetext{Pre-print. Under review.}}

\vspace{0.5em}

\section{Introduction}

Data-driven approaches to language technology have shifted the realm of possibility in multilingual NLP.
Distributed word representations \citep{mikolov2013distributed} have lifted the reliance on language-specific hand-crafted rules.
This is leveraged by pre-training language models on multiple languages simultaneously, such as multilingual BERT \citep{devlin-etal-2019-bert} and XLM-R \citep{conneau-etal-2020-unsupervised}, increasing performance through transfer learning.
More recently, even English-centric large language models (LLMs) are claimed to ``possess multilingual capabilities that surpass our expectations'' \citep{yuan2023multilingual}, in the case of LLama \citep{touvron2023llama}, and ``effectively transfer learned knowledge across different languages'' \citep{zhang-etal-2023-dont}, in the case of GPT3.5.

A shared factor behind the success of these models is their reliance on large volumes of textual data.
These data-driven approaches are claimed to be language-agnostic, as they are, in principle, applicable to any language given enough training data.
However, language-agnostic systems are not language-independent \citep{bender-2009-linguistically, bender2011achieving}.
Current algorithms are designed with an English-centric perspective in mind, while characteristics (such as low morphological complexity) of the English language cannot be assumed to transfer to other languages.
To {\textcolor{major-change-v3}{\sout{be}the}} best of our knowledge, there is currently no systematic investigation of how the properties of languages included in the evaluation influence the performance estimations of multilingual language models.
It remains unclear {\textcolor{major-change-v3}{\sout{how multilingual these models truly are and to what extent the language-independent assumption really holds against the world's language diversity.} how well these models perform on diverse languages systematically sampled.}}

To assess whether a language model performs well \textit{across} languages, ideally, one would evaluate it on all languages in the world.
However, collecting high-quality data on such a scale is not feasible.
Therefore, multilingual models are evaluated on a sample of the world's languages.
To {\textcolor{major-change-v3}{\sout{ensure}increase}} generalizability, such a language sample should be diverse, with varying characteristics and properties.
However, existing approaches to this sampling process have not been optimal.
Previous work has established that there is no clear terminology or methodological consensus for what constitutes `typological diversity' \citep{ploeger2024typological}.
Currently, many approaches in NLP use phylogenetic heuristics to ensure `typological diversity'.
In this paper, we show that this approximation of typological diversity through language phylogeny has severe shortcomings, and we provide a framework for systematically selecting languages based on typological distance measures.
Our framework implements two sampling methods that are widely established within the field of linguistic typology.
We demonstrate how our framework can be used for diverse typological language sampling, how it can help guide dataset expansions, and how it has use beyond typology.

\paragraph{Contributions} (i) We establish that language phylogeny is limited when it comes to assessing typological diversity; (ii) we provide a method for quantifying the typological distance between pairs of languages; (iii) we provide a systematic framework for selecting typologically diverse languages for multilingual evaluation scenarios; (iv) we introduce {\textcolor{major-change-v3}{four}} measures of typological diversity{\textcolor{major-change-v3}
; (v) we show that typological diversity matters in downstream evaluation; (vi) we provide a Python package that facilitates typologically diverse language selection, which is publicly available in the following repository:\\
\url{https://github.com/esther2000/typdiv-sampling}

\section{Background}
\label{sec:background}
{\textcolor{major-change-v3}{Representative language sampling has long been a central methodological issue in the field of linguistic typology. In order to assess its applicability to sampling in NLP, we discuss this line of research in more detail.}}

Linguistic typology can be described as the study of structural similarities and differences across the world's languages \citep{kashyap2019language}.
Within linguistic typology, an important research direction is the investigation of general patterns for these similarities and differences.
For example, \citet{greenberg1963universals} finds that in languages with verb-object word order, adpositions tend to be placed before their objects.
For drawing such general conclusions about human languages as a whole, testing linguistic hypotheses on only a handful of related languages is insufficient \citep{rijkhoff1993method,guzman2022statistical}.
Instead, generalizable conclusions in typology require adequate sampling strategies: findings should be supported by evidence across a diverse range of languages.
{\textcolor{major-change-v3}{\sout{Thus, representative language sampling has long been a central methodological issue in the field of linguistic typology.} Thus, language sampling has received considerable attention in typological research. }}
Given the recent advances in language technology beyond English, language sampling has become increasingly relevant for multilingual NLP.
Drawing generalizable conclusions about {\textcolor{major-change-v3}{\sout{multilingual model performance}the performance of multilingual models} requires tests across diverse languages.
Moreover, gaining insight into the skews of evaluation language sets may help to identify weaknesses of current applications.
In this section, we discuss relevant sampling strategies and terminology from the field of linguistic typology and how they relate to language sampling in NLP{\textcolor{major-change-v3}{, despite differences in objectives.}}.

\paragraph{Universe, Frame and Sample}
\citet{bell1978language} introduced central notions for language sampling in typology.
{\textcolor{major-change-v3}{To establish a common ground for approaching language sampling more systematically in NLP, we relate these terms to NLP.}}
Firstly, the sampling \textsc{universe} is the class of objects under study. In typological studies, this could for instance be the set of all possible human languages. In NLP, the sampling universe often corresponds to all existing natural languages that a certain language technology is claimed to generalize to.
The sampling \textsc{frame} provides access to the sampling universe. It is the concrete set of languages one can actually sample from.
\citet{sjoberg2023knowledge} further describes the distinction between a \textit{catalogue} frame, which is the set of languages we know to exist, and a \textit{corpus} frame, which is the set of languages for {\textcolor{major-change-v3}{which}} we have relevant research materials. For our purposes, we will treat the catalogue frame as all languages we have typological information for. We will treat the corpus frame as the set of languages that we have available datasets for.
Given the data requirements of most techniques in NLP, the corpus frame is of vital importance.
Lastly, the \textsc{sample} is the set of languages that one draws from the frame, with the aim to reflect the properties of the sampling universe. 
In NLP, this can correspond to the concrete set of languages that one tests or evaluates an application on.

\begin{figure}[h]
    \centering
    \includegraphics[width=0.5\linewidth]{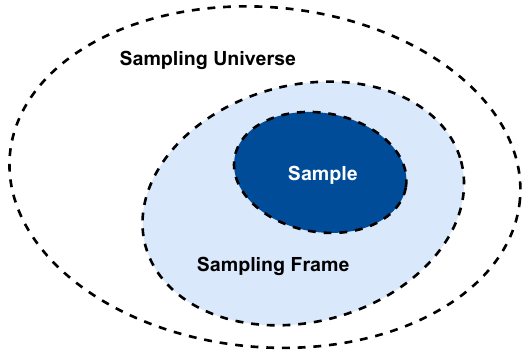}
    \caption{\textcolor{major-change}{A diagram showing the relationship between the sampling universe, sampling frame and the sample. The objective of diverse language sampling is to draw a sample from the sampling frame that optimally reflects a larger population (frame or universe).}}
    \label{fig:bell-diagram}
\end{figure}

\paragraph{Sampling Methods}
For extracting a sample from the frame, there are multiple popular types of methods in linguistic typology.
\citet{rijkhoff1998language} divide sampling methods into three categories: \textsc{random}, \textsc{probability} and \textsc{variety} sampling. Each of these sampling methods {\textcolor{major-change-v3}{\sout{are} is}} relevant for different research questions.
Random sampling entails selecting languages without any criteria. Without `stratification', the grouping of the frame before sampling, it is possible that resulting samples are in large part made up of similar languages.
As such, these samples could be skewed towards languages from certain phylogenetic or geographical groups.
This is especially likely if the sampling frame contains a skew.
Such samples, given they are large enough, are commonly used to look into the occurrence frequency of some linguistic phenomenon, but not necessarily for generalizable conclusions about language.
In probability sampling, one aims for a sample that contains independent languages. Ideally, the resulting sample should be free of bias. For instance, it should not contain a skew towards one or a few language families.
Variety sampling aims at sampling languages such that the linguistic diversity of the world's languages is captured as much as possible. This is because `exceptional types test the rule' \citep{perkins1988covariation}.
{\textcolor{major-change-v3}{It should be noted, however, that rather than creating statistical models of all languages of the world, sampling in NLP is rather concerned with testing multilingual model robustness across languages.}}
In the context of NLP, 
\citet{ponti-etal-2020-xcopa} write that choosing a variety sample tests the robustness of a language model to unseen typological features, as it includes outliers.
These three sampling methods each come with different implications in terms of sample size.
Because of the lack of stratification, a random sample `must be relatively large in size to be able to produce reliable results' \citep{rijkhoff1998language}.
For variety sampling, it is also often the case that the more languages are included, the better. This is because including more languages means that outliers and uncommon properties are more likely to be captured \citep{miestamo2016}.
However, for probability sampling, there is a trade-off between independence and coverage. The more languages one includes, the more difficult it becomes to preserve independence between these languages. To illustrate: if the number of languages one samples is larger than the number of language families in the frame, then multiple languages from the same family have to be sampled.

\paragraph{Language Sampling in NLP}
\citet{bender-2009-linguistically,bender2011achieving} and \citet{pikuliak-simko-2022-average} extensively discussed the need for diverse evaluation language selection, and the potential for linguistic typology to facilitate this.
However, to date, there are only a handful of works in NLP that aimed to apply these suggestions.
In fact, \citet{ploeger2024typological} ascertained that `typologically diverse' sampling methods in NLP are mostly flawed.
Firstly, if any stratification criterion is mentioned in NLP, this is usually based on phylogeny \citep[e.g.,][]{yadavalli-etal-2023-slabert,acs-etal-2021-subword,xu-etal-2020-modeling}.
To the best of our knowledge, no approach in NLP uses a tree structure to apply genealogical stratification. Instead, sampling from different families is common.
To illustrate: \citet{majewska-etal-2020-manual} `sampled languages from 5 different language families to ensure typological diversity'.
However, there is no evidence that phylogenetic relations directly imply typological relations \citep{dahl2008}. We further explore this in Section \ref{sec:phylogeny}.
Secondly, most sampling stratification is applied post-hoc. Commonly, a set of languages is selected, and only then is it described in terms of diverse language families and sometimes typological diversity.
For example, the typology index by \citet{ponti-etal-2020-xcopa} (see Section \ref{sec:metrics}) is only applied after language selection. As such, it does not promote principled language selection.
Our work differs from these approaches, in that we perform informed \textit{selection}, instead of {\textcolor{major-change-v3}{only}} informed \textit{analysis}.

\section{Why is Phylogeny Insufficient?}
\label{sec:phylogeny}
In Section \ref{sec:background}, we mentioned how previous work in NLP typically approximates typological diversity through phylogenetic groupings. More specifically, a major share of approaches claim to approximate typological diversity by randomly selecting languages from different language families or genera.
However, it is not evident that phylogenic relationships directly imply typological similarities \citep{georgi-etal-2010-comparing}.
Furthermore, relying on this stratification criterion can give vastly inconsistent samples (and thus different results), given that within-family selection is random.
In this section, we critically assess the shortcomings of phylogeny as a proxy for typologically diverse language sampling in NLP.

\paragraph{Theoretical Arguments}
Phylogenetic groupings such as language families and genera are commonly described as strictly exclusive groups.
For example, German, Dutch and Hindi are all in the Indo-European language family.
In reality, language similarity is much more gradient. 
Intuitively, it may be clear for some that German and Dutch are much more similar than Dutch and Hindi.
It is therefore not surprising that the strict boundaries between families are not necessarily agreed upon.
For example, \citet{dixon1997rise} writes ``about 1,000 languages have been grouped together in a putative `Niger-Congo family'. [...] One searches in vain for proof of this `genetic relationship.' ''
Furthermore, \citet{dahl2008} writes that ``[genealogically] related languages that are no longer in contact with each other can in a few thousand years develop typological profiles that are no longer indicative of a common origin.''
The imposed strict distinction between phylogenetic language groups also causes other issues.
For instance, pidgins and Creoles are not easily placed into a family tree, which means that sometimes they are excluded from sampling methods. For instance \citet{sjoberg2023knowledge} writes: ``Finally, pidgins and creoles are also excluded from the frame, due to the difficulty of deciding where in and in what family tree to place them.''
Furthermore, strictly divided language families come in different sizes. The Niger-Congo language family includes more than 1,000 languages \citep{glottolog}, while isolates such as Basque constitute their entire `family'. This further complicates fair sampling.
Finally, if the number of languages one wishes to sample is smaller than the number of phylogenetic groupings, there is a random selection on a group-level as well.

\begin{figure*}[h]
    \centering
\includegraphics[width=\textwidth]{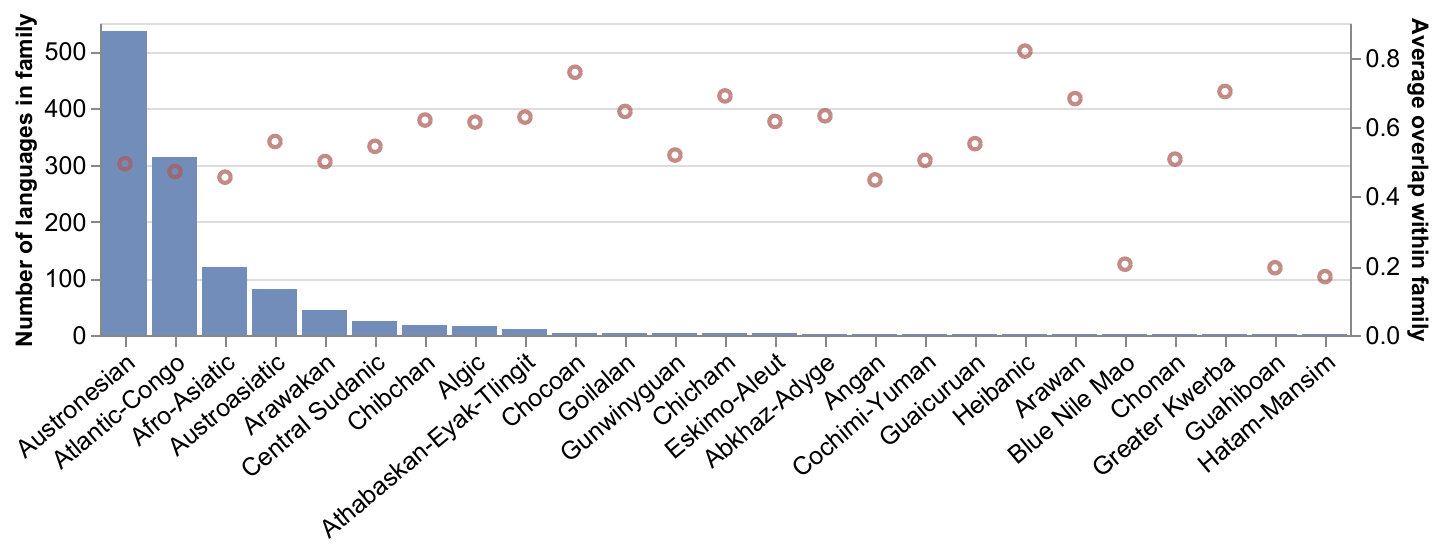}
    \caption{Number of languages and average within-family feature value overlap for the 25 largest language families in Grambank.}
    \label{fig:fam} 
\end{figure*}

\paragraph{Empirical Arguments}
Beyond theoretical reasons, it is unclear to what extent phylogenetic groupings actually imply typological diversity.
Here, we empirically assess the extent to which typological similarity overlaps with language families.
Grambank \citep{skirgaard2023dataset} contains genealogical and typological information for 2,467 languages.
We use this data to calculate, for each language family in Grambank, the pairwise average of overlapping feature values in that family. 
In Figure \ref{fig:fam}, we show this overlap per language family, as well as the number of languages in each family.
If families were coherent typological groups, we would expect to see high overlap throughout -- we observe variety in overlap, with most averages below 0.6. 
Additionally, we retrieve the closest language for each language in Grambank, in terms of feature value overlap. We find that the closest language is in another family in 32.42\% of the cases. Thus, sampling from distinct language families does not directly imply that the sampled languages are typologically distant. 

All in all, we conclude that sampling with phylogenetic stratification is not ideal for NLP purposes.
In linguistic typology, sampling using a proxy is often necessary. 
Directly using typological values instead is to be avoided, because then using those same variables for sampling introduces circularity.
However, in NLP, the variable under investigation is commonly something different, such as {\textcolor{major-change-v3}{\sout{multilingual model performance}the performance of multilingual models}. Thus, there is no circularity with sampling directly from typological features.
This avoids many of the issues with strict group sampling: we can take into account granularity, include languages which are typically difficult to place in one of those strict groups, we can control the group size and are not affected by randomness in sampling.
Additionally, by sampling through typological features directly, we gain immediate insight into the typological diversity of the sample.
In the next Section \ref{sec:framework}, we present a method that does exactly this.

\section{Related Work}
We argue that typologically generalizable evaluation in multilingual NLP should be conducted on the basis of \emph{a priori} sampling with typological features.
To the best of our knowledge, this is only implemented in two previous works.
\citet{dahl2008} sampled a subset of the World Atlas of Language Structures (WALS; \citealt{dryer2013wals}) by removing one language from each pair that is above a certain typological similarity threshold.
The language that is removed, is the one with the least coverage in WALS, which exacerbates bibliographical bias \citep{bakker2010language}.
\citet{stoll2013capturing} introduce a sampling method based on fuzzy clustering \citep{kaufman2009finding}. They divide languages into $k$ clusters based on twelve typological features, which the authors manually coded from various sources. Then, for each cluster, they sample the language with the highest membership coefficient.

Our work differs from these approaches considerably. Firstly, our method is designed to handle data-inequality. This means that it does not assume that all typological features are described for every language, contrary to previous work.
As complete typological coverage is rare, this renders our method applicable to a far broader range of languages and typological features.
Also, we present a flexible framework, where individual features can be left out, if desired.
Importantly, our framework accommodates both probability sampling and variety sampling from linguistic typology. This is important, because different sampling methods may be used to answer different research questions.

\section{A Principled Language Sampling Framework} 
\label{sec:framework}

The task of language sampling consists of selecting a set of languages (sample) from a larger set of languages (frame).
There are two main methods for performing  typologically diverse sampling.
For variety sampling, the languages in the sample should be maximally diverse. For probability sampling, these languages should be maximally independent.
Instead of approximating typological diversity through phylogenetic relationships, we measure typological distance between languages based on typological properties. Our approach consists of three steps:
\begin{enumerate}
    \item Retrieve typological information per language (Section \ref{sec:method-data})
    \item Calculate pairwise distances between languages (Section \ref{sec:method-distance})
    \item Sample languages using an algorithm that calculates a set of typologically distant languages (Section \ref{sec:method-algorithms})
\end{enumerate}
In Figure \ref{fig:framework}, we schematically represent our complete sampling pipeline. In the next subsections, we discuss each of the separate steps in more detail.

\begin{figure*}[h]
    \centering
    \includegraphics[width=0.95\textwidth]{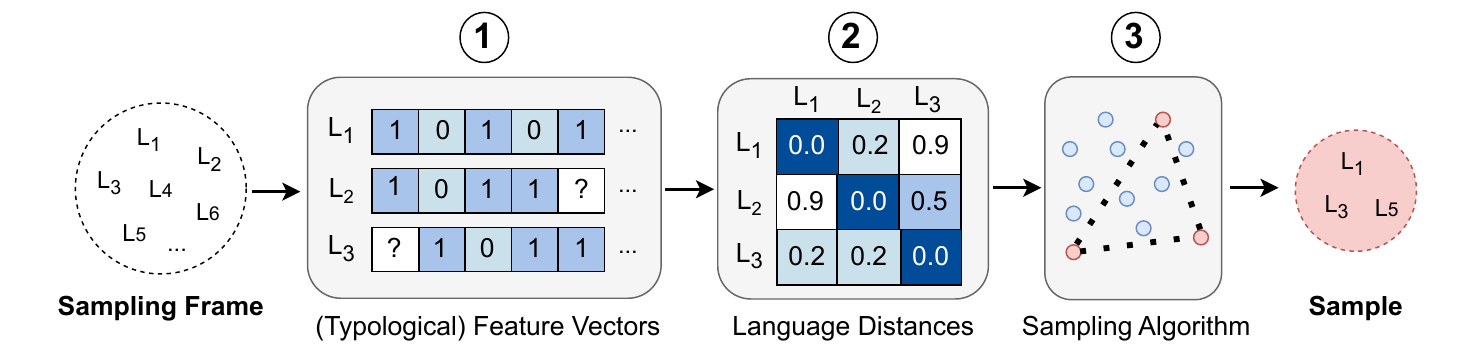}
    \caption{Language sampling by measuring distances of typological feature vectors.} 
    \label{fig:framework} 
\end{figure*}
\noindent

\subsection{Typological Feature Vectors}
\label{sec:method-data}
We use the Grambank database \citep{skirgaard2023grambank} for retrieving  typological characteristics of the languages in a sampling frame.
Grambank v1.0, which we use throughout this paper, contains typological information for 2,467 language varieties, for 195 grammatical features.
Of these features, 189 are expressed as binary statements (e.g., \textit{Are there definite or specific articles?}; GB020). These features can take values \textit{0}, \textit{1} or \texttt{?}, where the latter denotes unclear or unknown features.
Coverage is incomplete: for some language varieties, some features are not described, indicated by \texttt{no\_cov}.
Six word-order features are multi-value. For instance, feature GB024 (\textit{What is the order of numeral and noun in the NP?}) can take the values \textit{Num-N}, \textit{N-Num}, \textit{both} and \texttt{?}.

Our framework is compatible with any kind of information about languages.
The framework can also be used starting from step 3 when there are already pairwise distances available.
These language distances are not required to be based on typology.\footnote{We show an application of this in Section \ref{sec:geo-dist}.}
In this work, we specifically use Grambank, for multiple reasons.
Firstly, Grambank was developed with computational applications in mind, taking specific care to avoid logical dependencies between feature values \citep{haynie-etal-2023-grambanks}.
Logical dependencies are introduced when the value of one feature implies another. 
These are problematic for calculating (language) distances, as features that directly imply each other are then weighted more than features that are not logically implied by others.\footnote{It should be noted that a lack of \textit{logical} dependencies does not exclude statistical dependencies, as some feature values may co-occur.}
Additionally, since our work focuses on text processing specifically, the morphosyntactic domain of Grambank is particularly relevant.\textcolor{major-change-v4}{\st{By contrast, databases such as WALS also contain phonological features, while phonology is generally separate from text.}
} 
For our main experiments, we therefore treat Grambank as our catalogue frame.
Still, if one aims to conduct a phonological study, one could use phonological features with our framework instead.
{\textcolor{major-change-v3}{It should be noted that the representativeness of this sampling approach heavily depends on the input features.
For example, using input features that are only available for a few languages cannot retrieve representative samples more generally.
In addition, statistical dependencies between features can bias the sampling results, making it essential to relate the input features to the use case at hand.
}}

\subsection{Language Distance Calculation}
\label{sec:method-distance}

Let a sampling frame be a set $\mathcal{L}$ of languages. For each language $l \in \mathcal{L}$ we extract a typological vector $V(l)$ with $d$ dimensions, which consists of all Grambank feature values for $l$.
We binarize the six multi-value word order features, as suggested by \citet{haynie-etal-2023-grambanks}.
This leaves us with $d=209$ features which for each language will have values 0, 1, $\texttt{?}$ or $\texttt{no\_cov}$, the latter representing a feature not covered.
We treat both features without coverage and features with a value of $\texttt{?}$ as explicitly missing values, which we both treat as \texttt{NaN}.
For a vector $v$ and integer $i$, let $v_i$ be the $i$th feature of $v$.
For each pairwise combination of languages $l, l' \in \mathcal{L}$, we then calculate the euclidean distance in the presence of missing values, as defined by \citet{dixon} and implemented by \citet{scikit-learn}.
This distance calculation is defined in Equation \ref{eq:eucl}.

\begin{equation}
    \mathit{dist}(l,l') = \sqrt{w(l,l') \cdot \sum_{f\in \mathit{s}(l,l')}^{\;} ( V(l)_f - V(l')_f)^{2}}
    \label{eq:eucl}
\end{equation}
where $s(l,l')$ is the set of features that are covered in both $l$ and $l'$ 
\begin{equation}
    \mathit{s(l,l')} = \{f \in \{1\,..\,d\} \, | \,  V(l)_f \neq \texttt{NaN} \text{ and } V(l')_f \neq  \texttt{NaN}  \}
\end{equation}
and where $w$ is calculated by  dividing the total number of data points by the number of present data points:
\begin{equation}
    w(l,l') = \frac{d}{|s(l,l')|}.
\end{equation}

\noindent
This provides us with the distance between all pairs of language varieties in Grambank. We use these distances in our algorithms for calculating sets of typologically distant languages (Section \ref{sec:method-algorithms}).
Our framework provides additional flexibility by supporting a number of processing steps:
\begin{itemize}
    \item \textbf{Normalization:} For better readability of the pairwise distances, min-max normalization can be applied, retrieving distances in the range $[0,1]$. This does not affect the sampling process or results.
    \item \textbf{Binarization:} For using Grambank's multistate values in the same format as the binary features, we incorporate a binarization option.
    We follow the authors of Grambank by dividing each  multistate value into two binary features, which are not logically dependent.\footnote{\url{https://github.com/grambank/grambank/blob/master/docs/Grambank_most_updated_sheet.tsv}}
    \item \textbf{Language cropping:} In order to mitigate influences of languages with very limited feature coverage on the sampling results, we provide the option of removing languages with a defined percentage of missing data. Following \citet{skirgaard2023grambank}, we use a threshold of >25\% missing data for language cropping in the rest of this work.
    \item \textbf{Removing macro-languages:} Grambank contains a number of macro-languages (e.g. Central pacific linkage, Oceanic), which may not be relevant to one's case study. Our framework facilitates filtering these out, informed by the number of child languages in Glottolog.
    \item \textbf{Feature sub-selection:} For case studies that only comprise a subset of morphosyntactic features, our framework supports the selective inclusion of features.
\end{itemize}

For the demonstrated applications of our framework in this work, we apply normalization, binarization, language cropping and remove macro-languages.

\subsection{Sampling Algorithms}
\label{sec:method-algorithms}

Given these typological distances between pairs of languages, there are multiple ways to do typologically diverse language selection.
In the linguistic typology literature, different sampling methods are used for answering different kinds of research questions. 
Inspired by this, our framework explicitly accommodates both variety sampling and probability sampling methods.
While we specifically focus on typology in this section, these methods can be used with any type of information, as mentioned in Section \ref{sec:method-data}.

\subsubsection{MaxSum Diversity}

For our typologically informed \textit{variety} sampling, we treat the problem as a \textsc{maximum diversity problem (MDP)}.\footnote{\citet{marti2013heuristics} list a number of alternative names used for the problem: maximum dispersion, max-avg dispersion, $p$-dispersion, $p$-dispersion-sum, edge-weighted clique, remote clique, maximum edge-weighted subgraph, dense $k$-subgraph, $p$-defense, $p$-defence-sum and equitable dispersion.} 
This entails finding a size $k$ set of points where the sum of distances between all points in the set is maximal (\textit{MaxSum}). 
Approaching the problem as such optimizes for sampling for large \textit{total} distance, capturing outliers (see Figure \ref{fig:mdp_mmdp_comparison}), which is the objective of variety sampling.

\citet{kuo1993analyzing} showed that the `clique problem', finding subsets of vertices in a graph that are all adjacent, can be reduced to MDP. Since the clique problem is NP-hard, then so must MDP be. Famously, there is no known efficient algorithm for finding optimal solutions to NP-hard problems.
We also experienced that a brute force algorithm did not terminate when run on the dataset. \citet{marti2013heuristics} give an overview of heuristics for MDP and conclude that even simple heuristics give good solutions. We implement one such simple heuristic.
Given pairwise language distances, we first find the  language that is most distant from all others. For this, we take the language where the sum of distances with all other languages is largest. The motivation behind this is that selecting the language with the most `unusual' typological properties (or combinations thereof) already ensures some outlier.
Next, we add the language that is furthest from the first language to the sample. We then sum the distances from these two languages to every other language. The language that has the largest summed distance is added to the selection.
We repeat this process until the desired size of the sample is reached. The full procedure is described in Algorithm \ref{algorithm_maxsum}.

\subsubsection{MaxMin Diversity}
In sampling with the MaxSum objective, the total typological distance between languages is maximized. Individual outliers within the sample may be close (see Figure \ref{fig:mdp_mmdp_comparison}), which may introduce a skew in a sample.\footnote{
The distribution of language types can be visualized as distributed normally around the center, see \citet{dryer1998statistical}.}
However, for \textit{probability} sampling, one would want to preserve the independence between languages.
To this end, we approach the diversity problem as a \textsc{MaxMin Diversity Problem}, which is a second popular diversity sampling objective \citep{parreno2021measuring}. 
Instead of optimizing for the maximum total distance, we optimize for maximizing the smallest distance between any two points in the set. Thus, we select languages such that the closest two are maximally typologically distant.
This problem is NP hard as well, which means we again need to rely on heuristics.
Similar to MaxSum diversity, we implement a simple heuristic.
As in the MDP approach, we first select the language that is most distant from all others. Then, we again take the language furthest from that language.
Then until the desired sample size ($k$) is reached, we add the next language to the sample that retrieves the highest minimum distance to the already selected points.
This last step is the key difference between the objectives of both algorithms.
The full process is described in Algorithm \ref{algorithm_maxmin}.

\algrenewcommand\algorithmicindent{0.5em}%
{\centering
\begin{framed}
\textbf{Input:} $k$: number of languages to sample, $\mathcal{L}$: sampling frame, $\mathit{dist}$: function giving the pairwise distance between languages in $\mathcal{L}$ (e.g., Equation \ref{eq:eucl}) \\
\begin{minipage}{0.48\textwidth}
\begin{algorithm}[H]
   \begin{algorithmic}[1]
    
        \State $l \gets \argmax_{l \in \mathcal{L}} \sum_{l' \in \mathcal{L}} \mathit{dist}(l, l')$
        \State $L \gets \{ l \}$
        \While{$|L| < k$}
            \State $L \gets L \; \cup $\newline
                \hspace*{2em}$\{ \argmax_{l \in \mathcal{L} \setminus L} \sum_{l' \in L} \mathit{dist}(l, l') \}$
        \EndWhile
        \State \textbf{return} $L$
        \newline
   \end{algorithmic}
    \centering
    \caption{MaxSum Sampling }\label{algorithm_maxsum}
\end{algorithm}
\end{minipage}
\hfill
\begin{minipage}{0.48\textwidth}
\begin{algorithm}[H]
    \begin{algorithmic}[1]
        \State $l \gets \argmax_{l \in \mathcal{L}} \sum_{l' \in \mathcal{L}} \mathit{dist}(l, l')$
        \State $l' \gets \argmax_{l' \in \mathcal{L}} \mathit{dist}(l, l')$
        \State $L \gets \{ l, l' \}$
        \While{$|L| < k$}
            \State $L \gets L \; \cup $ %
            \newline
                \hspace*{1em}$\{ \argmax_{l \in \mathcal{L} \setminus L}(\min_{l' \in L} \mathit{dist}(l, l')) \}$
        \EndWhile
        \State \textbf{return} $L$
    \end{algorithmic}
    \centering
    \caption{MaxMin Sampling}\label{algorithm_maxmin}
\end{algorithm}
\end{minipage}
\end{framed}
} 

\noindent
\textcolor{major-change}{Within linguistic typology, variety sampling is used to test a hypothesis `against the extremes', while probability sampling is used for research questions that require more even language distributions. In Section \ref{sec:div-gen}, we further contrast these methods for downstream use cases in NLP, showing that MaxSum sampling is especially useful for scenarios in which \emph{a worst-case performance estimate is desired}.}

 \begin{figure}[H]
    \centering
    \includegraphics[width=0.45\textwidth]{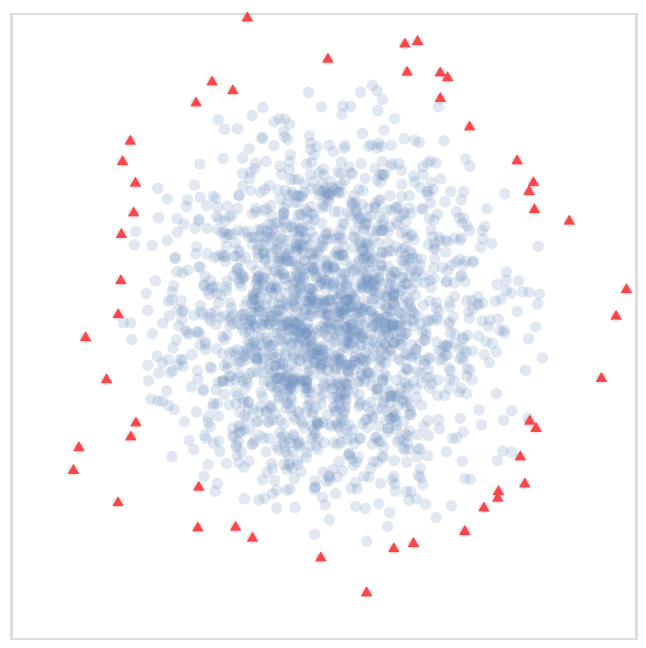}
    \hspace{2em}
    \includegraphics[width=0.45\textwidth]{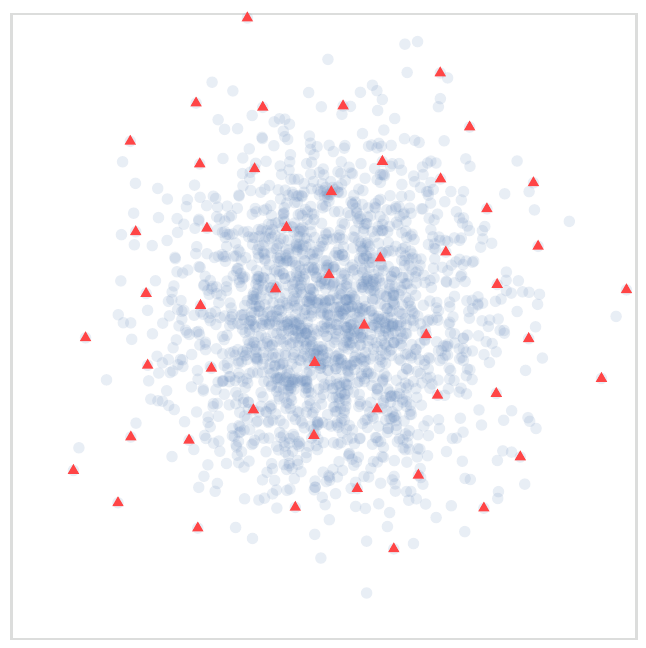}
    \caption{A visualization of both sampling algorithms, with the MaxSum objective on the left and MaxMin on the right over a normal distribution. The \textcolor{algored}{red} triangles represent the sample selected by the respective algorithm, the \textcolor{algoblue}{blue} dots the remaining languages in frame. The distance here is the Euclidean distance in a 2D plane.}
    \label{fig:mdp_mmdp_comparison}
\end{figure}

\section{Typological Diversity Evaluation}
\label{sec:typ-eval}

Our sampling methods enable a priori language sampling for typological diversity. In Section \ref{sec:use_cases} we look into how our algorithm can be used practically.
Here, we first verify whether our sampling methods actually retrieve more diverse language samples than previous methods used in NLP.
To this end we compare our sampling algorithms to approaches from previous work in multilingual NLP with four metrics.
Additionally, we address the issue of how many languages one would need to include for a representative sample.

\subsection{Baselines}\label{sec:baselines}

\paragraph{Random} As mentioned in Section \ref{sec:background}, language selection in NLP is mostly unprincipled, where evaluation languages are sampled without stratification methods. This resembles random sampling in linguistic typology. 
In typology, random samples are mostly used for gaining insight into occurrence frequencies, but not for drawing generalizable conclusions.
For comparability, we include a baseline which randomly samples from the sampling frame (Grambank).

\paragraph{Convenience} In practice, random sampling in NLP is not truly random, because data availability plays a large role.
Therefore we select the languages that are most commonly used in previous work in NLP that claim to have `typologically diverse' language selections.
\citet{ploeger2024typological} annotated the language selections in papers containing such claims.
For our baseline, we sort these languages based on occurrence frequency, and sample the first $k$.
In case of a tie in terms of occurrence frequency, selection is random.

\paragraph{Phylogenetic}
The most popular stratification in NLP, if any, is sampling based on language families.
This method can be motivated from the perspective of linguistic typology, where phylogenetic stratification is popular. Contrary to linguistic typology, advanced sampling methods from language trees are not relevant to NLP (Section \ref{sec:background}).
We implement phylogenetic stratification by sampling uniformly from groupings on two levels: language families\footnote{As specified in Glottolog v4.4: 215 families.} and genera\footnote{As specified in WALS v2020.3: 612 genera.}.
If the requested sample size is bigger than the number of groupings, we uniformly sample again from the groupings until we reach the requested size.
Note that this {\textcolor{major-change-v3}{is}} already more principled than most sampling approaches in NLP.
The convenience baseline is indirectly also often influenced or motivated by phylogeny, albeit in less principled manner.
As such, our {\textcolor{major-change-v3}{\sout{phylogenentic}}phylogenetic}  baseline is a `best case scenario' for principled, typologically diverse language sampling using phylogenetic data.  

\subsection{Metrics}
\label{sec:metrics}
Measuring the typological diversity of a language set can be done in a number of ways.
We formulate and compare four diversity metrics here, incorporating both previous work and new ideas.
We indicate `higher is better' metrics with an up arrow ($\uparrow$) and `lower is better' with a down arrow ($\downarrow$).

\begin{figure}
    \centering
\includegraphics[width=\textwidth]{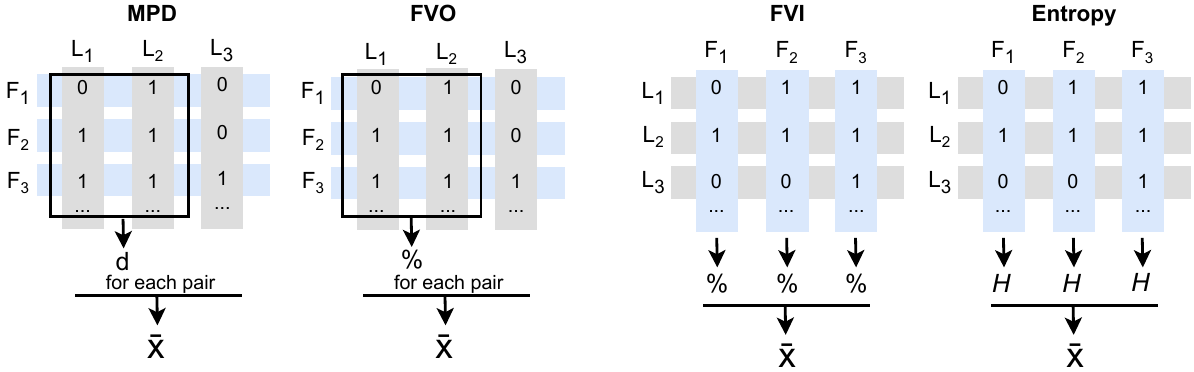}
    \caption{Schematic overview of  diversity metric calculations, where `for each pair' means all combinations of languages in the set.}
    \label{fig:metrics-schematic}
\end{figure}

\paragraph{Mean Pairwise Distance (MPD)}
The mean pairwise distance (MPD) of a language set measures the euclidean distance between all combinations of languages in the sample \citep{ploeger2024typological}.
Since these distances are used in our sampling algorithms directly, this measure serves as a sanity check to verify whether the distances that the algorithm is based on are actually increased.
Pairwise comparisons can be motivated from a typological perspective \citep{wichmann2010pairwise}, and taking the average means that the results can be compared across language samples of different sizes. MPD can be formalized as follows:
\begin{equation}
    \mathit{MPD}(L) = \frac{1}{|L|(|L|-1)} \sum_{l, l' \in L, l \neq l'} \mathit{dist}(l, l')
\end{equation}
Here $\mathit{dist}(l, l')$ denotes the euclidean distance as defined in Equation \ref{eq:eucl}.

\paragraph{Feature Value Overlap (FVO)}
Distances alone do not directly describe the disparity of our features.
This is motivated from the perspective of linguistic typology; \citet{dahl2008} describes measuring language similarity as: ``How large a proportion of the features that are defined for both members of a language pair have different values?''.
To this end we calculate the feature value overlap (FVO), which is the average of the percentages of features that overlap between any pair of languages in the combinations of a language set.
Since Grambank contains binarized feature values (as outlined in Section \ref{sec:method-distance}), calculating such an overlap is appropriate.\footnote{Special care has to be taken to calculate this metric when using multistate features values.}
We report the average over all combinations.
Feature value overlap can be formalized as follows:
\begin{equation}
    \mathit{FVO}(L) = \frac{1}{|L|(|L|-1)} \sum_{l, l' \in L, l \neq l'}\frac{|\{f \in \{1\,..\,d\} |\, V(l)_f = V(l')_f  \}|}{d}
\end{equation}

\paragraph{Feature Value Inclusion (FVI)}
The previous metrics do not measure the extent to which individual typological properties are covered. This is especially relevant for variety sampling, where rare typological features should be included.
\citet{miestamo2016} defined the measure of \textit{saturation} of a typological feature as ``the proportion of values, out of the maximum number of possible values, found in the sample for that feature''.
\textcolor{major-change}{As the name suggests, this metric is expected to saturate for larger language selections. Therefore, it is especially suitable for comparing smaller language samples.}
Because we deal with binary features only, we calculate the feature value inclusion (FVI) per feature as the percentage of languages that include the feature. We report the average over all features. FVI can be formalized as follows:

\begin{equation}
    \mathit{FVI}(L) = \frac{1}{d} \sum_{f=1}^{d} \frac{|\{V(l)_f \, | \, V(l)_f \neq \texttt{NaN}   \text{ and } l \in L \}|}{2}
\end{equation}

\paragraph{Entropy ($H$)}
Similar to the diversity index reported in \citet{ponti-etal-2020-xcopa}, we report the entropy of the feature values that occur in a language sample. 
This gives us insight into the spread within features.
For example, if a sample $L_{1}$ contains $[1,1,1,1,0]$ for a given feature $f$, FVI is maximal, but the spread is low.
The entropy for $f$ is lower than {\textcolor{major-change-v3}{that of}} a more diverse sample $L_{2}$ with values $[1,1,0,0,0]$.
We take the average of this metric over all features.
A difference with previous work is that \citet{ponti-etal-2020-xcopa} based their entropy calculations {\textcolor{major-change-v3}{\sout{based}}}on 103 unnamed typological features from URIEL \citep{littell-etal-2017-uriel}. This is not reproducible, as they do not report which typological features were used. Furthermore, as URIEL contains logical dependencies, the included typological features are not necessarily weighted equally.
The entropy of a set of languages is the average entropy over all features:
\begin{equation}
    H(L) = \frac{1}{d} \sum_{f=1}^{d} H(f)
\end{equation}

where the entropy of a feature is

\begin{equation}
    H(f) =  -\sum_{i \in \{0,1\}} p(f,i) \cdot \log_2 p(f,i)
\end{equation}

and the probability $p$ is calculated as

\begin{equation}
    p(f,i) = \frac{|\{l \in L\, |\, V(l)_f = i  \}| }{ | \{l \in L \, | \, V(l)_f \neq \texttt{NaN})\} | }.
\end{equation}

\subsection{Results and Discussion}
\label{sec:results}
We compare our sampling methods against baselines based on previously used methods in NLP, as introduced in Section \ref{sec:baselines}.


Figure \ref{fig:metrics-by-k} shows that our methods consistently retrieve more diverse samples than all baselines.
For the pairwise metrics (MDP, FVO), this difference is especially large for smaller samples; it is easier to avoid overlap when sampling fewer languages.
This is in line with earlier findings from typology, which described the trade-off between coverage and independence \citep{miestamo2016}.
While the FVI difference across methods is small for large sample sizes, we find that our methods retrieve a higher inclusion of feature values for small samples ($< 20$) than the baselines. This is especially relevant for NLP, where multilingual hypotheses are commonly tested on a small set of `typologically diverse' languages (Median = 11; \citealt{ploeger2024typological}).
Lastly, we observe that MaxSum and MaxMin consistently retrieve samples with higher feature entropy than other methods.

\begin{figure}[h]
    \includegraphics[width=\textwidth]{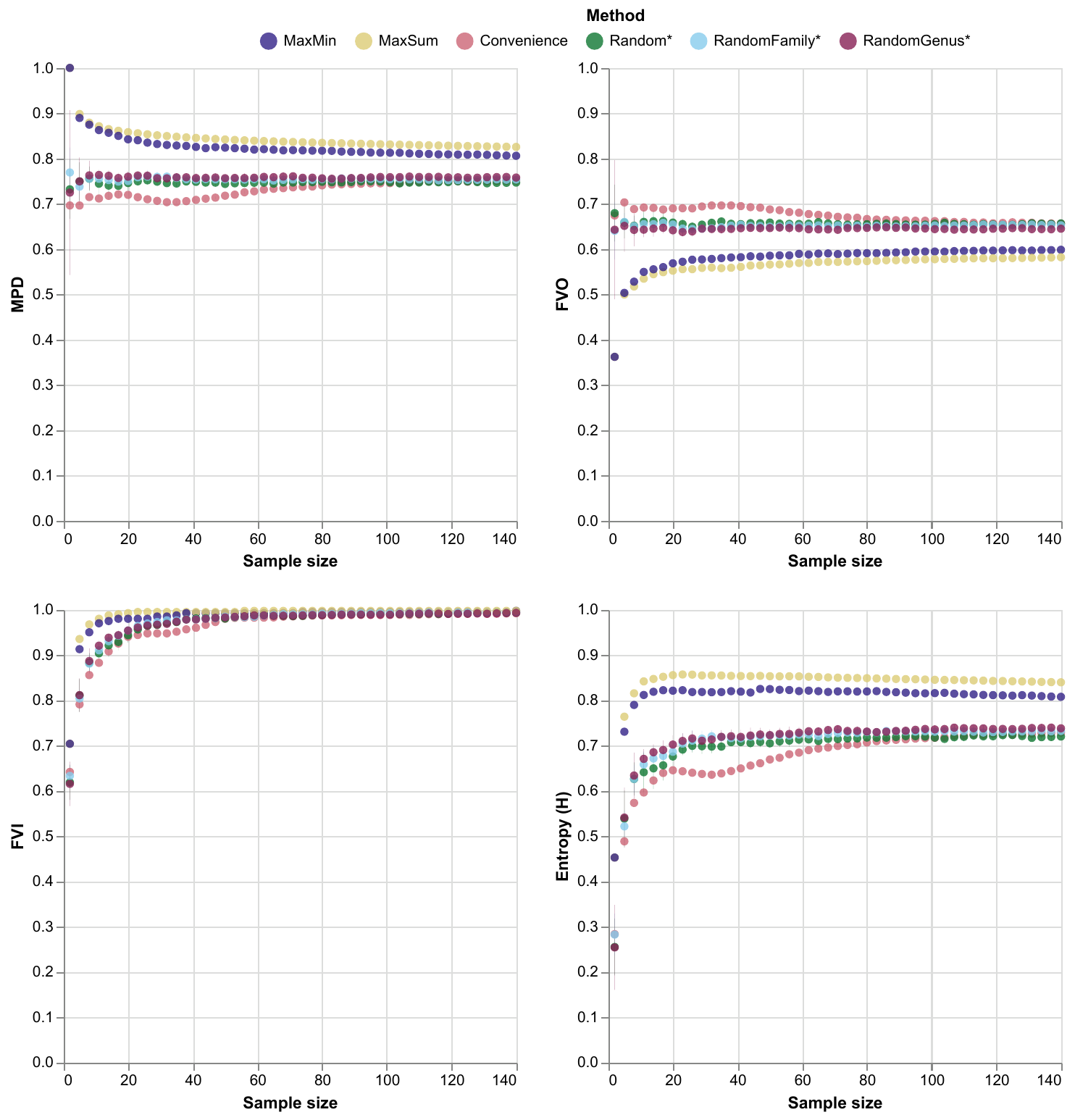}
    \caption{MPD ($\uparrow$), FVO ($\downarrow$), FVI ($\uparrow$) and $H$ ($\uparrow$) for different sample sizes.
    Non-deterministic methods are indicated with an asterisk and averaged over 10 runs, error bars represent their standard deviation.}
    \label{fig:metrics-by-k}
\end{figure}

These results also showcase how our methods can be used to inform the choice of sample size in the design of a study.
For these particular corpus frames, it can be argued that a sample size of 20 languages adequately covers the feature space. For instance, FVI flattens after $k > 20$.
This could be a justification for making claims about the generalizability of a certain phenomenon captured in a sample.\footnote{While the number of languages one can sample from is a characteristic of the method, we find similar results when taking the intersection of all methods' sampling frames: see Appendix \hyperref[app:intersection]{A}.}

\noindent
Next, we compare the methods {\textcolor{major-change-v3}{\sout{more in detail} in more detail}}, for a specific value of $k$. We zoom in on $k=20$, since the metrics tend to flatten off for all methods around that value, as seen in Figure \ref{fig:metrics-by-k}.
We measure the MPD, FVO, FVI and $H$ for the samples that all our sampling approaches retrieve. We run the non-deterministic methods 10 times.
The results (Table \ref{tab:intrinsic_eval}) show that both MaxSum and MaxMin sampling retrieve considerably better results than all baselines.

\begin{table}[]
    \centering
    \caption{Results for $k = 20$ with all languages in Grambank as the frame, where an asterisk indicates methods that are non-deterministic, these are averaged and the standard deviation over 10 random runs is listed.}
    \scalebox{0.85}{
    \begin{tabular}{lllll}
	\toprule
	\textbf{Sampling Method} & \textbf{MPD} $\uparrow$           & \textbf{FVO} $\downarrow$         & \textbf{FVI} $\uparrow$           & $\boldsymbol{H}$ $\uparrow$ \\\midrule
	Convenience*             & 0.72 ${\scriptscriptstyle\pm0.00}$ & 0.69 ${\scriptscriptstyle\pm0.00}$ & 0.94 ${\scriptscriptstyle\pm0.00}$ & 0.65 ${\scriptscriptstyle\pm0.00}$     \\
	Random*                  & 0.75 ${\scriptscriptstyle\pm0.01}$ & 0.66 ${\scriptscriptstyle\pm0.02}$ & 0.94 ${\scriptscriptstyle\pm0.02}$ & 0.68 ${\scriptscriptstyle\pm0.02}$     \\
	RandomFamily*            & 0.75 ${\scriptscriptstyle\pm0.01}$ & 0.65 ${\scriptscriptstyle\pm0.01}$ & 0.95 ${\scriptscriptstyle\pm0.01}$ & 0.69 ${\scriptscriptstyle\pm0.02}$     \\
	RandomGenus*             & 0.76 ${\scriptscriptstyle\pm0.02}$ & 0.64 ${\scriptscriptstyle\pm0.01}$ & 0.95 ${\scriptscriptstyle\pm0.01}$ & 0.70 ${\scriptscriptstyle\pm0.02}$     \\\midrule
	MaxSum                   & \textbf{0.86}                   & \textbf{0.55}                   & \textbf{0.99}                           & \textbf{0.86}                                \\
	MaxMin                   & 0.84                            & 0.57                            & 0.98                   & 0.82                      \\
    \bottomrule
    \end{tabular}
    }
    \label{tab:intrinsic_eval}
\end{table}

\clearpage
\section{\textcolor{major-change}{How does Typological Diversity relate to Generalizability?}}
\label{sec:div-gen}

\textcolor{major-change}{
In this work, we argue that increasing the linguistic representativeness of a language sample leads to better generalizability of results in multilingual NLP.
Currently, multilingual language technology is typically only evaluated on a handful of seemingly randomly selected languages.
This lack of systematic language sampling makes it difficult to assess how multilingually generalizable such technology really is.
In this section, we assess whether sampling typologically diverse languages actually leads to better estimates of general multilingual performance.
We apply our sampling framework to 
downstream NLP tasks, demonstrating how sampling diverse test languages affects the generalizability of the results.
Specifically, we look into natural language understanding (topic classification), natural language generation (machine translation) and subword segmentation.}
{\textcolor{major-change-v3}{Note that, as illustrated in Figure \ref{fig:bell-diagram}, the objective of systematic language sampling is to draw a language sample from the frame that optimally reflects the larger population.
Thus, we evaluate the extent to which the distribution of results in typologically diverse samples reflects the a distribution over a large group of languages.
}}
}

\textcolor{major-change}{\subsection{Natural Language Understanding}}

\noindent
\textcolor{major-change}{To assess the extent to which samples drawn with different methods reflect general multilingual performance, it is important {\textcolor{major-change-v3}{\sout{we} to}} sample from a dataset that includes sufficient typological variation.
The Taxi1500 dataset \citep{taxi1500}, which spans more than 1500 languages\footnote{The intersection between Grambank and Taxi1500 contains 1506 languages.}, covers a large part of the `typological space' (see Appendix \hyperref[app:frame-div]{B}). Importantly, its evaluation set is parallel across languages, which facilitates cross-lingual comparison.
The benchmarking task of Taxi1500 is classifying the topic of bible verses, where the six possible topics are \textit{recommendation}, \textit{faith}, \textit{description}, \textit{sin}, \textit{grace} and \textit{violence}.
These topics were manually annotated on an English bible, and applied to other languages through annotation projection to create a large-scale parallel benchmark.
We compare the results of widely popular classification models (mBERT, \citealp{devlin-etal-2019-bert}; XLM-R, \citealp{conneau-etal-2020-unsupervised}) as well as the more recently introduced Glot500-m \citep{imanigooghari-etal-2023-glot500}.
}

\textcolor{major-change}{
First, in Table \ref{tab:taxi1500}, we compare the average scores over the samples, and their similarity to the average over all languages.
Ideally, diverse language sampling should lead to averages that are more comparable to the average over all languages.
We observe that this is indeed the case with our sampling methods, and that especially results obtained with the convenience baseline provide overly optimistic pictures of general multilingual performance, as compared to the more realistic performance estimates obtained with our carefully sampled language selections.
Importantly, we highlight differences between MaxSum and MaxMin: MaxSum gives more conservative performance estimates than MaxMin, since, similar to variety sampling, it samples the most extreme cases.
This may be useful to NLP practitioners, especially in scenarios where a worst-case estimate is desired.
}

\textcolor{major-change}{
In Figure \ref{fig:taxi1500-results}, we go beyond average results, and inspect the \textit{distribution} of the results.
We compare the distribution of the accuracy over all 1506 languages with samples ($k=20$) from two all methods: MaxSum, Maxmin (ours) and convenience (reflecting common practice in NLP). We observe that the convenience baseline gives consistent overestimations of multilingual performance, while the results of our systematic sampling method consistently reflect the distributions of results for all languages in the dataset.
{\textcolor{major-change-v3}{\sout{Note that, as illustrated in Figure \ref{fig:bell-diagram}, the objective of systematic language sampling is to draw a language sample from the sampling frame that optimally reflects the larger population.}}}
This implies that performance estimates on true typologically diverse languages indeed paint a more generalizable picture of {\textcolor{major-change-v3}{\sout{multilingual model performance} the performance of multilingual models}}.
}

\begin{table}[th]
    \centering
    \caption{\textcolor{major-change}{Average results (accuracy) per sampling method ($k=20$) on Taxi1500 dataset. For the non-deterministic sampling method, we report the average over 10 runs.}}
    \scalebox{0.92}{
    \begin{tabular}{l|rrr|rrr|r}
    \toprule
         & \multicolumn{3}{c|}{\textit{\textbf{Non-deterministic Methods}}} & \multicolumn{3}{c|}{\textit{\textbf{Deterministic Methods}}} \\
         & \bf Rand. & \bf Rand. Fam & \bf Rand. Gen. & \bf Conv. & \bf MaxSum & \bf MaxMin & \bf All \\
     \midrule
         \bf mBERT & 14.3 ${\scriptscriptstyle\pm1.66}$ & 12.7 ${\scriptscriptstyle\pm1.22}$ & 13.5 ${\scriptscriptstyle\pm1.31}$ & 39.7 & 12.7 & 13.7 & 13.7 \\
         \bf XLM-R-B & 13.0 ${\scriptscriptstyle\pm2.91}$ & 9.7 ${\scriptscriptstyle\pm1.91}$ & 11.0 ${\scriptscriptstyle\pm2.81}$ & 56.8  & 7.7 & 13.8 & 11.0\\
         \bf XLM-R-L & 12.4 ${\scriptscriptstyle\pm3.34}$ & 8.6 ${\scriptscriptstyle\pm2.05}$& 10.3 ${\scriptscriptstyle\pm3.14}$ & 59.7 & 5.9 & 13.6 & 10.1 \\
         \bf Glot500-m & 16.8 ${\scriptscriptstyle\pm3.64}$ & 10.8 ${\scriptscriptstyle\pm1.70}$& 12.9 ${\scriptscriptstyle\pm3.01}$ & 54.6 & 9.6 & 13.9 & 14.4 \\
     \bottomrule
    \end{tabular}
    }
\label{tab:taxi1500}
\end{table}

 \begin{figure}[th]
    \centering
\includegraphics[height=15.2em]{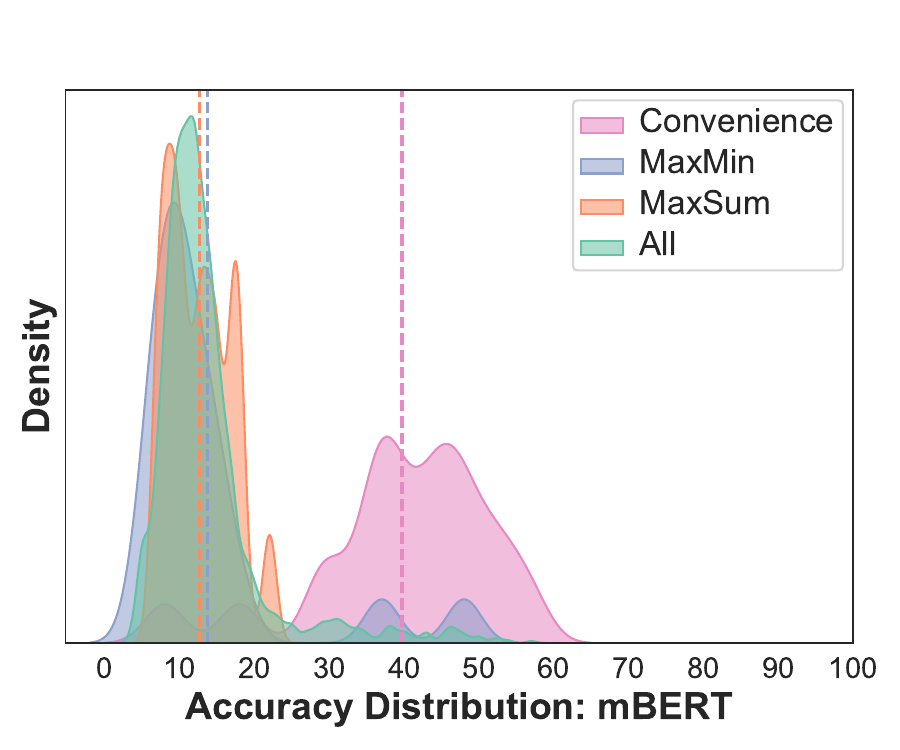}
    \hspace{1em}
\includegraphics[height=15.2em]{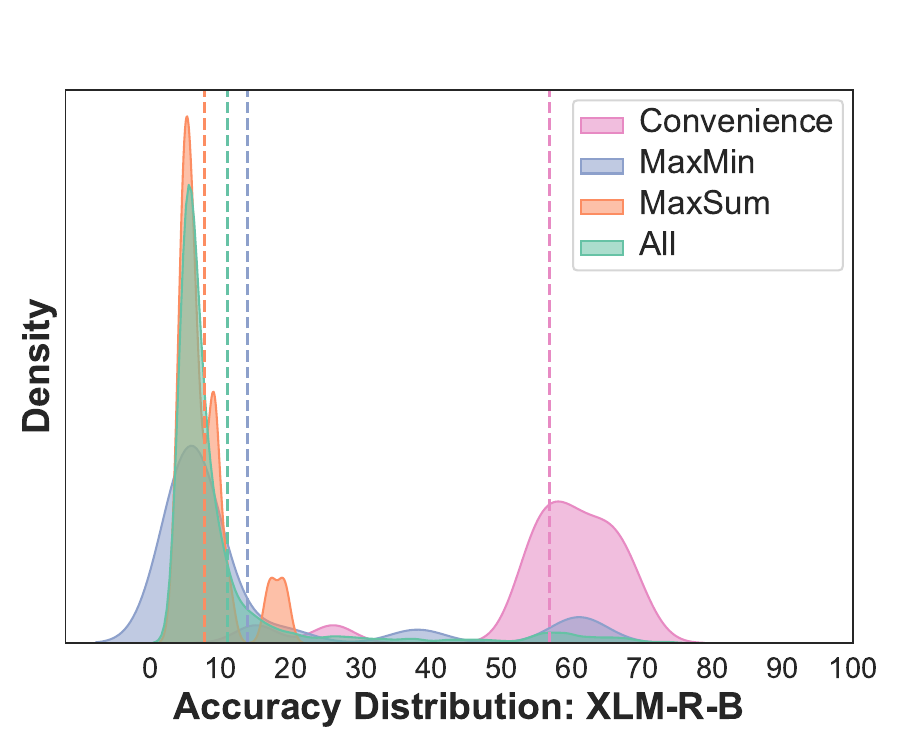}
    \hspace{1em}
\includegraphics[height=15.2em]{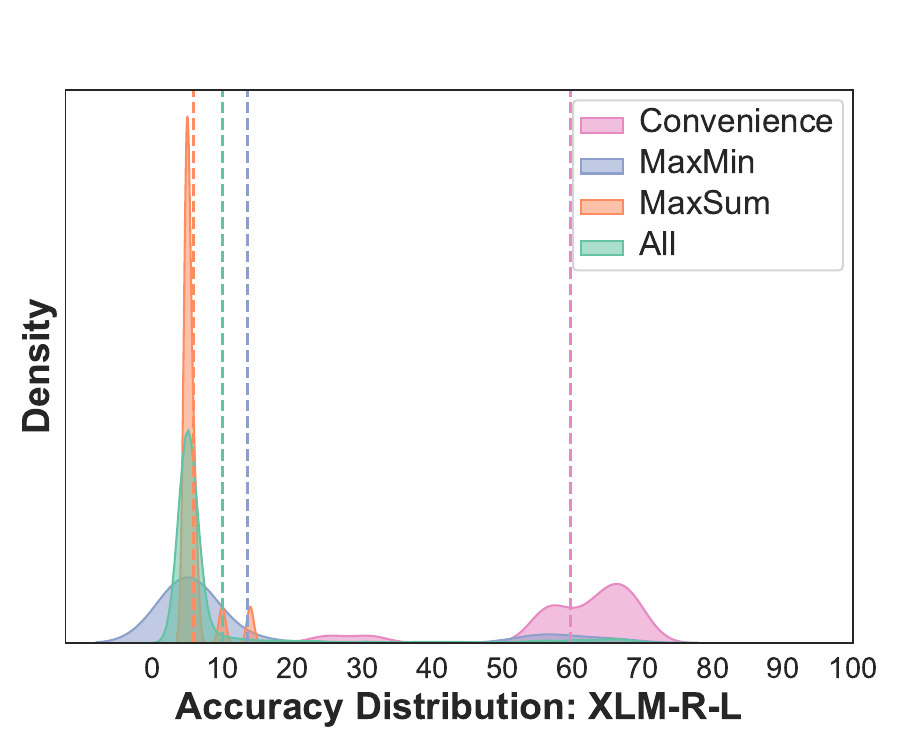}
    \hspace{1em}
\includegraphics[height=15.2em]{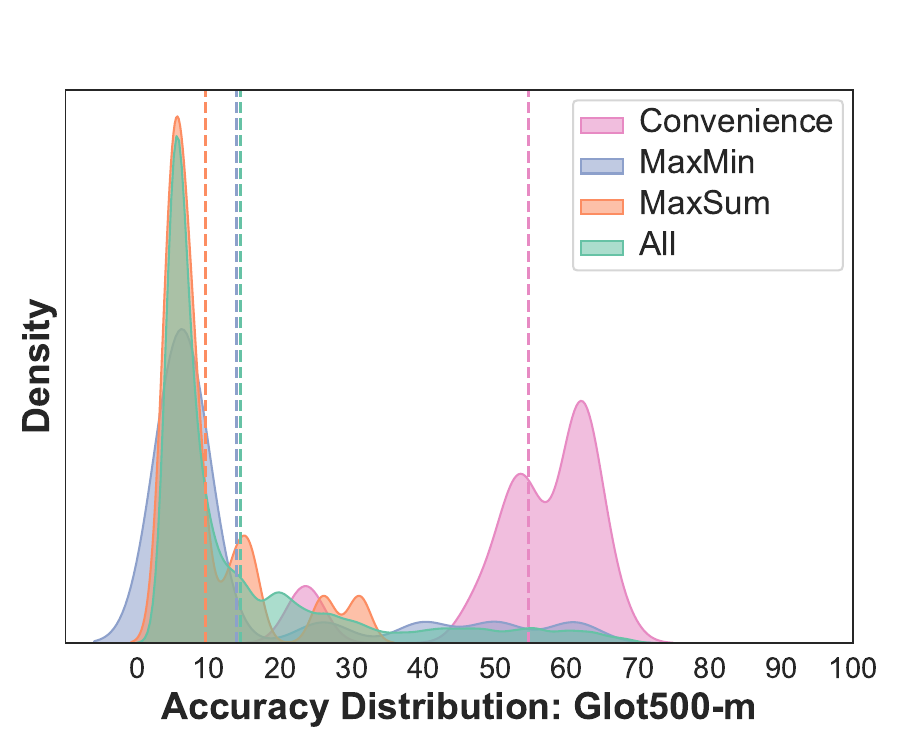}
\caption{\textcolor{major-change}{Accuracy \textcolor{major-change-v3}{of zero-shot performance} distributions of language samples ($k=20$) on the Taxi1500 evaluation set, contrasting all four deterministic sampling methods. Vertical lines represent averages per sampling method. Convenience sampling consistently over-estimates performance, as compared to MaxSum and MaxMin.}}\label{fig:taxi1500-results}
\end{figure}

\textcolor{major-change-v3}{Of course, inclusion or exclusion of languages in the training data for these models can be expected to affect the results. In Appendix \hyperref[app:train-lang-res]{C}, we show the results when sampling only from the pre-training languages of each model. The same pattern holds: the convenience baseline systematically overestimates multilingual performance. Note that in our analysis, we investigated the zero-shot performance of the multilingual models, thus fine-tuning languages are not a confounding factor.}

\textcolor{major-change}{\subsection{Natural Language Generation}}

\noindent
{\textcolor{major-change}{Beyond natural language understanding, we evaluate the effect of diverse sampling on a generative task, namely machine translation (MT).
We evaluate the effect of the source language on MT performance to English, using the multilingual NLLB-200 model (3.3B parameters).\footnote{\url{huggingface/facebook/nllb-200-3.3B}}
The results are retrieved from the OPUS Dashboard \citep{tiedemann-de-gibert-2023-opus}.
Due to cross-lingual differences, MT performance can only be compared on the same target language and the same test sets.
We choose to focus on investigating translation to English specifically, as this is the only language for which the dashboard contains results from all source languages in FloRes-200.
English-centric MT has widely been prioritized in MT research, largely motivated by speaker numbers and information access, as argued by \citet{bugliarello-etal-2020-easier}, among others.
Furthermore, \textcolor{major-change-v3}{\sout{to keep the focus on} since we do not propose a general language `leaderboard', but rather systematically investigate the influence of} typological characteristics {\textcolor{major-change-v3}{on NLP evaluation}}, we avoid the confounding factor of writing systems. We 
\textcolor{major-change-v3}{\sout{only take into account languages that are written in Latin script.}control for script in our experiments by sampling from languages that use the Latin script specifically, since this maximizes the number of languages in our sampling frame.
}
\textcolor{major-change-v4}{
It should be noted that this limits the generalizability of the findings of this experiment, since we exclude a sizable group of languages.
}
}}
\begin{table}[h]
    \centering
        \caption{\textcolor{major-change}{Average results (BLEU) per sampling method ($k=20$) on FloRes-200 dataset. Non-deterministic sampling methods are run 10 times.}}
    \scalebox{0.85}{
    \begin{tabular}{ccc|ccc|c}
    \toprule
         \multicolumn{3}{c|}{\textit{\textbf{Non-deterministic Methods}}} & \multicolumn{3}{c|}{\textit{\textbf{Deterministic Methods}}} \\
         \bf Rand. & \bf Rand. Fam & \bf Rand. Gen. & \bf Conv. & \bf MaxSum & \bf MaxMin & \bf All \\
     \midrule
           34.74${\scriptscriptstyle\pm3.29}$ & 32.74 ${\scriptscriptstyle\pm1.18}$ &33.00 ${\scriptscriptstyle\pm0.98}$ & 46.64 & 30.05 & 31.52 & 32.69\\
     \bottomrule
    \end{tabular}
    }
    \label{tab:mt}
\end{table}

\begin{figure}[h]
    \centering
    \includegraphics[width=0.7\linewidth]{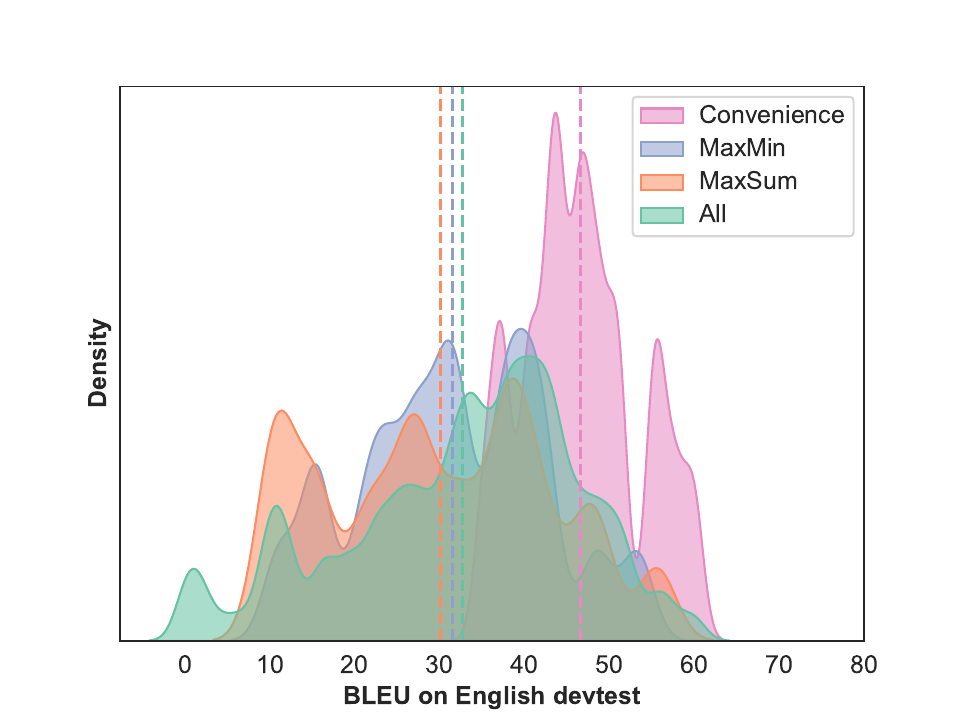}
    \caption{\textcolor{major-change}{Distribution of results (BLEU) of many-to-English MT with NLLB-3.3B, on the languages in Latin script from the FloRes-200 dataset. Vertical lines represent averages.}}
    \label{fig:xx-to-eng}
\end{figure}

The results, as shown in Table \ref{tab:mt} and Figure \ref{fig:xx-to-eng}, demonstrate that convenience sampling gives an overestimation of performance of many-to-English MT.
While performance may also be determined by the similarity to English, we see that
MaxMin and MaxSum retrieve averages and distributions that resemble the distribution over the entire sampling frame.
Furthermore, sampling with the MaxSum objective leads to a more conservative performance estimate than the MaxMin objective\textcolor{major-change-v3}{\sout{, consistently highlighting its potential for application in worst-case scenarios.} Thus, MaxSum sampling is more suitable for application in worst-case evaluation scenarios, whereas the average of MaxMin samples can be expected to better reflect a general distribution. }}\\

\subsection{Subword Segmentation}
Subword tokenization \citep{sennrich2016neural} is an important component in multilingual text processing, which is at the basis of many popular LLMs.
{\textcolor{major-change-v3}{While it is not a downstream task, suboptimal tokenization can have large implications for downstream results.}}
Splitting tokens into subwords facilitates better generalizability to languages with complex morphology, and better handles out-of-vocabulary tokens.
Yet, tokenizers of popular multilingual models have been shown to retrieve varying results across languages with varying morphosyntactic properties \citep{gutierrez-vasques-etal-2021-characters,gutierrez2023languages}. 
Highly synthetic languages such as Finnish pose different tokenization challenges than languages with relatively little morphological complexity such as Dutch. 
The number of subwords can indicate over-segmentation, which can have far-reaching consequences for a user: for instance, the token-based cost of LLM APIs is higher, and processing more separate tokens introduces latency \citep{petrov2024language}. 
{\textcolor{major-change-v3}{
Previous work \citep{ahia-etal-2023-languages,petrov2024language} therefore links segmentation to the concept of multilingual fairness.
To extend the scope of our evaluation beyond strict downstream results to fairness more generally, we include a systematic comparison of subword segmenters.
}}

Here, we conduct a large-scale cross-lingual sampling comparison of subword tokenization with tokenizers of popular multilingual models.
For comparability between languages, the data should ideally be parallel across languages. Previous work
\citep{ahia-etal-2023-languages,petrov2024language} evaluates on the FLoRes-200 dataset, as this is multi-parallel.
We considerably extend upon their language coverage in our analysis, by evaluating on text from the Parallel Bible Corpus \citep{mayer-cysouw-2014-creating}, which is the largest massively multi-parallel dataset in terms of language coverage.
We first match language ISO codes to Glottocodes. We then select all languages that have Grambank coverage, and control for script (i.e. we filter out non-Latin scripts). This retrieves 571 languages, which is the broadest language coverage thus far in tokenization analysis.
This is important, because we aim to test to what extent the sample represent{\textcolor{major-change-v3}{s}} the sampling frame, and potentially the sampling universe.
We select the longest bible per language, and sample 1,000  commonly available verses for our evaluation, to retain multi-parallelism.
We analyse tokenizers of four popular language models with tokenizers publicly available on HuggingFace\footnote{\url{https://huggingface.co}}: multilingual BERT \citep{devlin-etal-2019-bert}, \textcolor{major-change}{LLama 3.3 70B Instruct \citep{dubey2024llama}, Qwen2.5-32B \citep{qwen} and Phi 4 \citep{phi4technicalreport}.}
Similar to \citet{ahia-etal-2023-languages}, we measure the amount of segmentation through the average number of subwords per verse.\footnote{We do not use the fertility measure \citep{rust-etal-2021-good}, because it assumes similar word tokenizer performance across languages. We do not use `premiums', the disparity of tokenization length between parallel sentences in two languages \citep{petrov2024language}, because these are by definition relative to another language; we instead want to compare \textit{across} languages directly.}

 \begin{figure}[h]
    \centering
\includegraphics[height=15em]
{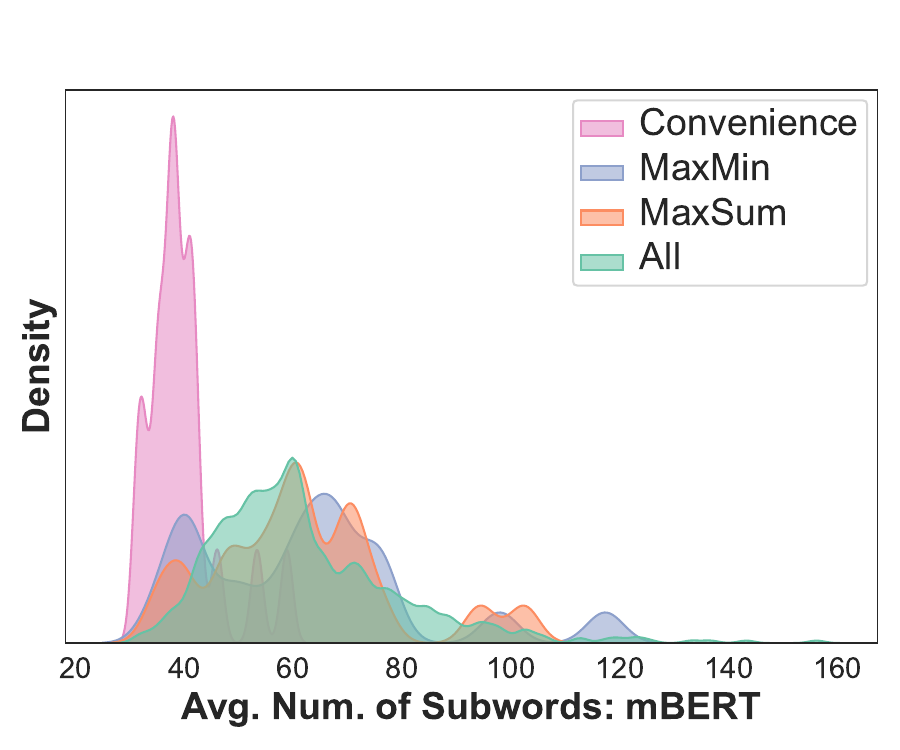}
    \hspace{1em}
\includegraphics[height=15em]{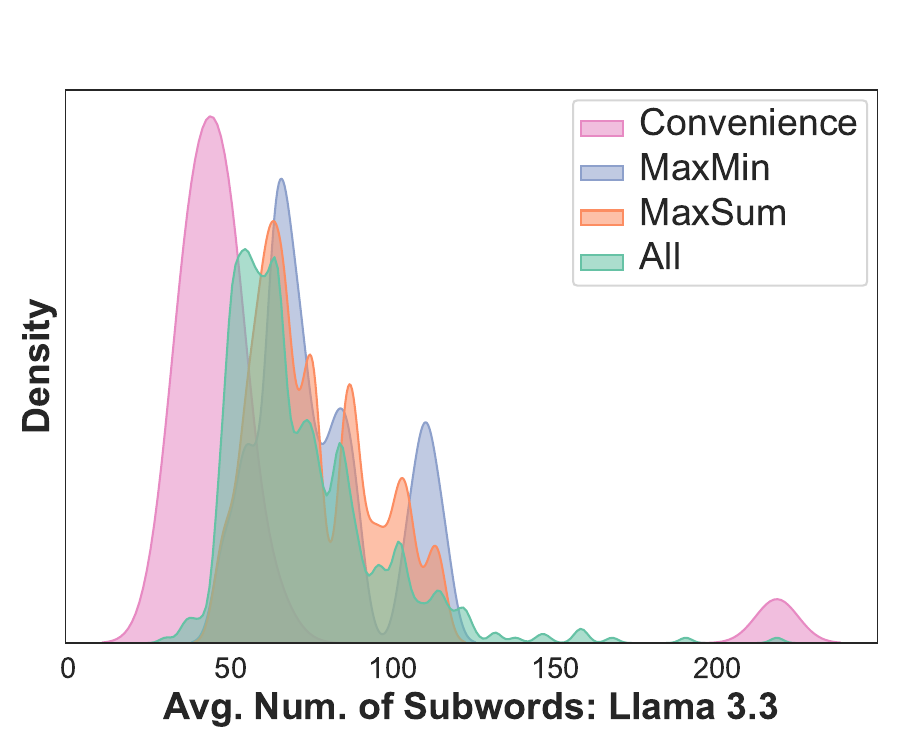}
    \hspace{1em}
\includegraphics[height=15em]{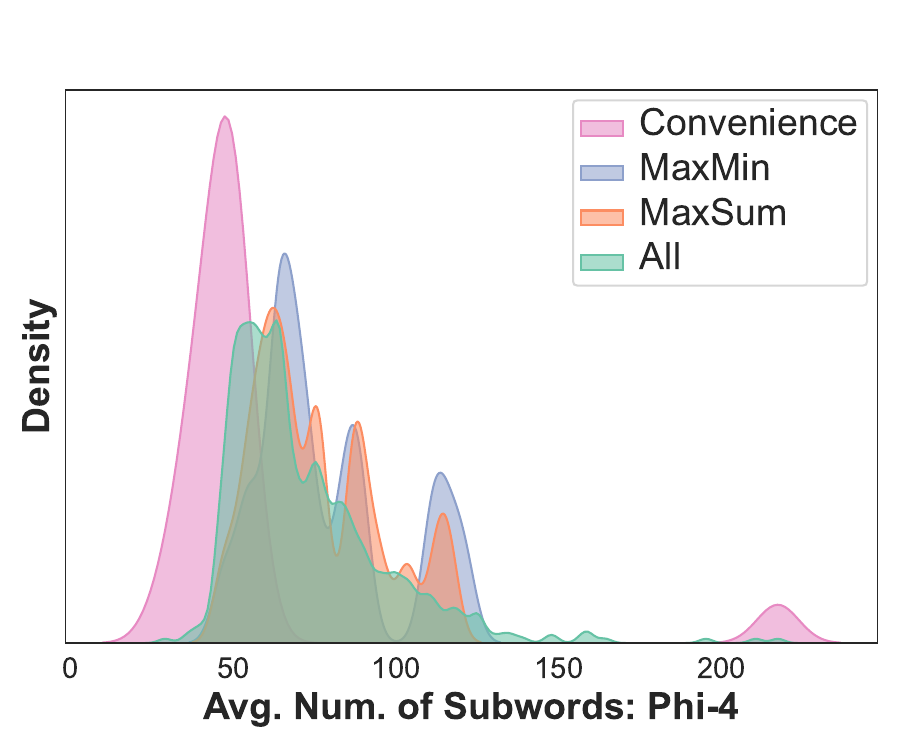} 
    \hspace{1em}
\includegraphics[height=15em]{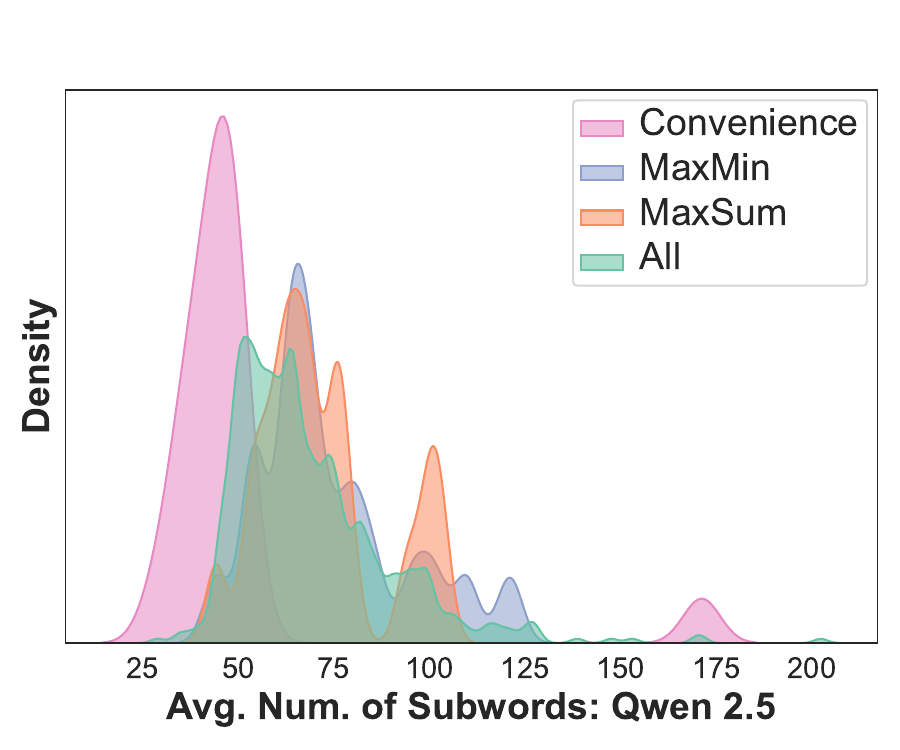} 

    \caption{Average number of subwords per verse across four popular tokenizers, with different sampling strategies.}
\label{fig:tokenization}
\end{figure}

We compare our sampling methods (MaxSum, MaxMin) with the only baseline that is deterministic for $k=20$: convenience.
From the 571 total languages, we sample 20 with each method.
For each tokenizer, we compare the spread in average number of subwords per verse for all three sampling methods, and compare those with the spread for all 570 languages in our experiment.
This gives us an estimate of how generalizable the sample of 20 languages is with respect to the total 500+ languages included in the experiment.

The results are {\textcolor{major-change-v3}{shown}} in Figure \ref{fig:tokenization}.
For all four models, we observe that the convenience baseline retrieves a considerably lower average number of subwords, with a generally smaller range than other methods.
This implies that evaluating on the 20 most commonly included languages in `typologically diverse' samples in NLP gives an overly optimistic image of \textit{general} multilingual tokenizer performance.
Yet, MaxMin retrieves averages and spreads that more closely resemble the average and spread of \textit{all} 500+ languages in the experiment.
This suggests that a priori typologically informed language sampling can improve the generalizability of the results, based on a language sample.

\section{Further Use Cases}
\label{sec:use_cases}

Our framework can be used in a variety of ways.
In this section, we provide examples of practical use cases, to inspire future research with systematic language selection.
Contrary to the experiments in Section \ref{sec:typ-eval}, we now also deal with a corpus frame.
This means that we do not merely sample from a typological database, but investigate more realistic data availability situations.
Our experiments, frames and samples are publicly available in our code repository.\footnote{\url{https://github.com/esther2000/typdiv-sampling/use_cases}}

\subsection{Guiding Dataset Expansion Efforts}
\label{sec:data-exp}

Data availability is an obstacle for truly diverse sampling, as corpus frames may be limited for certain NLP tasks.
At the same time, data collection and annotation efforts can be laborious and expensive.
Thus, it may be useful to know beforehand how data collection for one language may impact the generalizability of an evaluation set.
Our framework can be used to assess the diversity of existing benchmarks, inform future data collection efforts, and quantify the relative improvement in language diversity.

We present a use case with five popular existing multilingual evaluation benchmarks.
They represent a range of tasks, including: machine translation (FloRes-200; \citealt{nllb2022}), dependency~parsing~(Universal~Dependencies~v2.14; \citealt{zeman2024universal}), question answering (TyDiQA; \citealt{clark-etal-2020-tydi}), commonsense reasoning (XCOPA; \citealt{ponti-etal-2020-xcopa}) and generative language modelling (Aya Evaluation Suite (human-annotated); \citealt{singh2024aya}).
We focus on human-curated datasets specifically, as human annotations are typically more costly to gather than automatically generated data, and thus informed expansion is more relevant.
The benchmarks vary in size and in the extent to which the included languages were carefully selected.
For example, for FloRes-200 and Universal Dependencies (UD), there seem to be no explicit selection criteria, as expanding language coverage was the main objective.
Yet, some other datasets were explicitly created with typological diversity in mind.
TyDiQA was created with the aim to include typologically diverse languages. However, in line with our findings in Section \ref{sec:background}, the authors do not specify systematic sampling criteria and only post-hoc mention typological features to ``highlight the breadth of phenomena'' of the included languages.
The authors of XCOPA explicitly aim for variety sampling, but only seem to measure the typological diversity of their sample post-hoc.

Here, we quantify the diversity of each benchmark dataset, and use our framework to assess which expansion language would most increase typological diversity, and how this impacts the total set.
To this end, we provide a starting sample, which is the intersection of the languages in the datasets, and those in Grambank.
For retrieving the next-best language, we then choose a sample size of the starting sample, plus one. Note that this number can be raised in future applications.
We sample with the MaxSum objective, as this is effective in increasing the total diversity (Section \ref{sec:results}).
We report diversity based on entropy ($H$) and feature value inclusion (FVI), as discussed in Section \ref{sec:metrics}).

\begin{table}[h]
    \centering
    \small
    \caption{The expansion language that most increases typological diversity for five popular multilingual benchmarks, as retrieved with MaxSum sampling.}
     \scalebox{0.96}{
    \begin{tabular}{lccc|c|cc}
    \toprule
       \textbf{Dataset} & $|\boldsymbol{L}|\;(|\boldsymbol{L}\,\cap\,\textbf{GB}|)$ & $\boldsymbol{H}$ & \textbf{FVI} & \textbf{+ Language} & $\boldsymbol{H}'$ & \textbf{FVI}$'$ \\ 
    \midrule
       FloRes-200 & 195 (105) & 0.717 & 0.988 & Tariana & 0.721 & 0.988 \\
       \citep{nllb2022} & & & &  &\\
   \midrule
       UD v2.14 \citep{zeman2024universal} & 158 (80) & 0.681 & 0.985 & Tariana & 0.687 & 0.985\\
  \midrule
       TyDiQA \citep{clark-etal-2020-tydi} & 11 (7) & 0.627 & 0.883 & Movima & 0.693 & 0.928\\
  \midrule
       XCOPA \citep{ponti-etal-2020-xcopa} & 11 (7)  & 0.599 & 0.873 & Tariana & 0.667 & 0.913 \\
  \midrule
       Aya Evaluation Suite (human- & 7 (5) & 0.571 & 0.841 & Yele & 0.660 & 0.898 \\
       annotated) \citep{singh2024aya} & & & &  &\\
  \bottomrule
    \end{tabular}
    }
    \label{tab:next_best}
\end{table}
\noindent

The results (Table \ref{tab:next_best}) show that adding the next-best languages to the existing datasets always increases total diversity in terms of entropy.
This increase is relatively small for the larger datasets (FloRes-200, UD), but larger for the smaller datasets (TyDiQA, XCOPA, Aya Evaluation Suite), where the number of added languages is a larger proportion of the total.
FVI is not increased by adding a single language to the larger datasets.
This is unsurprising, as the value was already nearly maximal.
This suggests that including as many languages as possible is a good strategy for maximizing the included features, as argued for in variety sampling \citep{miestamo2016}.

Interestingly, one language stands out as the most diverse expansion language for multiple datasets (FloRes-200, UD and XCOPA): Tariana (language family: Arawakan).
From a linguistic typology perspective, this makes sense. Firstly, Tariana is spoken in ``a very remote area'' in the Vaupés area in the North West of Brasil,  which is ``not easy to get to'' \citep{aikhenvald2003teaching}. 
Also, Tariana is a
polysynthetic language \citep{aikhenvald2003language}, a grammatical property that is likely uncommon in popular NLP datasets.
Yet, there are more extensive, cultural, reasons for Tariana standing out.
The area where the language is spoken{\textcolor{major-change-v3}{\sout{,}}} is characterized by ``obligatory multilingualism, dictated by the principles of linguistic exogamy'' \citep{aikhenvald2003language}. This means that marriage only occurs between speakers of different languages.
There has been a ``strong inhibition against ‘language-mixing’, viewed in terms of lexical loans.'', and the language includes ``independent innovations [...] divergent from those found in closely related languages.'' \citep{aikhenvald2003teaching}.
While this case is an indicator that our method indeed achieves high diversity and thus does what we expect, we do not argue that Tariana should then be included in all NLP datasets. Instead, speaker needs and data availability should be taken into account, which can be done by narrowing the sampling frame. We demonstrate such a case in the next section.

\paragraph{Case Study: Universal Dependencies}
So far we have used all (cropped) languages in Grambank as the sampling frame.
In practice, this may not be realistic, as one might want to take into account speaker needs and annotator availability.
The sampling frame can then be smaller.
To analyse our framework in such a scenario, we narrow the sampling frame for UD to the languages in their \textit{Possible Future Extensions} list.\footnote{As listed on the homepage at the time of writing: \url{https://universaldependencies.org}.}
These are ``[languages for which] people have expressed interest in providing annotated data [...], but [for which] no valid data has been provided so far''.
We manually annotate the Glottocodes for these languages, and find that the intersection of previously not included Glottocodes and Grambank contains 17 languages.
These languages are the sampling frame for finding the `next best' language. We apply MaxSum sampling, which retrieves as next best extension language: Seri.

\begin{table}[H]
    \centering
    \caption{Effects in diversity metrics from adding Seri to UD v2.14.}
    \begin{tabular}{ll|ll|ll|ll}
    \toprule
        \bf MPD $\uparrow$ & \bf MPD' $\uparrow$  & \bf FVO $\downarrow$ & \bf FVO' $\downarrow$ & \bf FVI $\uparrow$ & \bf FVI' $\uparrow$ &  $\boldsymbol{H}$ $\uparrow$ & $\boldsymbol{H}$'  $\uparrow$ \\
    \midrule
        0.725 & \underline{0.728} & 0.679 & \underline{0.677} & \underline{0.985} & \underline{0.985} & 0.681 & \underline{0.685} \\
    \bottomrule
    \end{tabular}
    \label{tab:ud_nextbest}
\end{table}
\noindent
Again, this makes sense from a linguistic typology perspective, as Seri is a language isolate which ``does indeed have some very special characteristics'' \citep{marlett2000seri}.
In Table \ref{tab:ud_nextbest}, we report the effect of adding Seri to UD v2.14 on all our metrics.
We observe that, {\textcolor{major-change-v3}{\sout{which} with}} the exception of FVI, diversity is improved with respect to all metrics.

\subsection{Other Distance Maximisation}\label{sec:geo-dist}
Beyond typological features, our method serves primarily as a \textit{framework}, which can be extended to other features that describe languages.
For example, it can be used to sample languages that are spoken in geographically distant areas.
Such language sampling has been an important methodological focus in linguistic typology.
Similar to sampling from genealogical groupings, geographically diverse sampling is commonly performed with pre-defined groupings, such as macroareas \citep{dryer1989large, dryer1992greenbergian, hammarstrom2014some} or the finer-grained AUTOTYP areas \citep{nichols2009autotyp}.
In Section \ref{sec:phylogeny}, we established that the usefulness of phylogenetic language groupings can be hindered by their lack of granularity.
The same applies to geographical language groupings.
Firstly, all languages from one such area (e.g. `Eurasia') are seen as being equally close.
Secondly, languages that are geographically close, but not in the same macroarea, are not seen as similar.
Most previous work in typology that uses absolute geographical distance instead of groupings \citep{jaeger2011mixed,cysouw2013chapter,bjerva-etal-2020-sigtyp} manually defines a set threshold, for instance of 1,000 kilometers.\footnote{Still, it should be noted that distance-based stratification does not take into account long-distance contact, and that reducing languages to coordinates is not necessarily accurate.}

\begin{figure}[H]
\centering
\includegraphics[width=\textwidth]{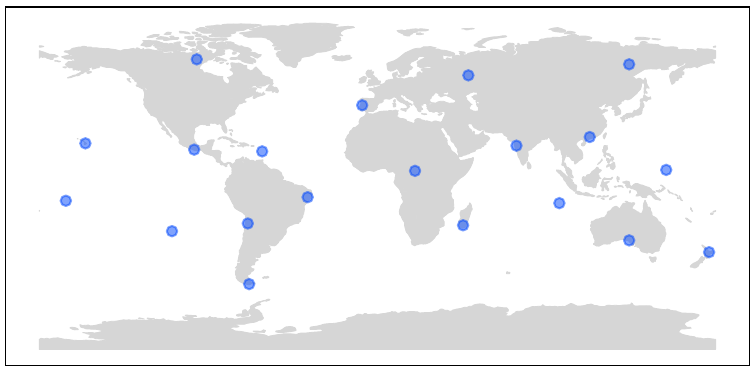}
\caption{We use MaxMin sampling to find twenty geographically distant languages, based on coordinate pairs.}
\label{fig:geo-dist}
\end{figure}

\noindent
Our framework allows for diverse geographical sampling, directly from distances, without the need to manually define a distance threshold.
We further demonstrate this use case here.
We first retrieve language coordinates from Grambank, which corresponds to step 1 in our framework. 
We use these to calculate the absolute distance in kilometers between any two coordinates (step 2).
These pairwise distances are then used for sampling geographically diverse languages.
Specifically, we use the MaxMin objective, with the objective to  maximize the minimum distance between any two points.
Figure \ref{fig:geo-dist} shows the resulting sample when selecting twenty languages.

\section{Limitations, Ethical Considerations and Discussion}

Our framework enables typologically diverse language sampling.
Critical assessment of language diversity is vital for the evaluation and development of multilingual language technology that is fair across languages.
Still, it should be noted that {\textcolor{major-change-v3}{\sout{generalizabilty} generalizability}} across typological characteristics constitutes only a fragment of multilingually fair NLP.
For instance, Grambank only addresses grammatical phenomena.
Such typological databases do not provide cultural information, which may be key for certain research questions \citep{hershcovich-etal-2022-challenges}.

Moreover, the feature coverage in typological databases such as Grambank is incomplete, which should be taken into account when drawing conclusions.
{\textcolor{major-change-v3}{As such, our evaluation is not completely representative. }}
However, since bibliographic bias in NLP research tends to be much stronger than in typological databases (see convenience baseline, Section \ref{sec:results}), we believe that information from typological databases can actually enable informed expansion of language coverage.
{\textcolor{major-change-v3}{For example, through informed dataset expansion (Section \ref{sec:data-exp}), we hope our framework can be used to further increase typological representativeness in future datasets.}}

Lastly, we acknowledge that reducing languages to coordinates or points in multidimensional space is by default simplistic.
Languages are more than objects of study: they are central to human communication and inherently involve humans.
As such, we urge researchers to consider factors beyond typological diversity in language sampling or dataset expansion.
Instead of sampling \textit{only} based on typological diversity, we emphasize the importance of incorporating a human-centered perspective.
While developing more generalizable multilingual NLP tools has the potential of mitigating unequal access to language technology, this should be a collaborative effort that involves speakers \citep{bird-2020-decolonising}.

\section{Conclusions}
In this work, we systematically analyse common language sampling strategies in NLP, and find that these are insufficient for typologically diverse language sampling.
Guided by research in linguistic typology, we propose two sampling algorithms.
We compare the samples obtained by our methods with strong baselines.
These samples are evaluated with four typological diversity metrics that show that our method consistently retrieves language samples with higher typological diversity.
{\textcolor{major-change}{
For downstream tasks, we show that our sampling algorithms provide a much more realistic estimate of generalizability as compared to, e.g., convenience sampling, thereby highlighting the importance of addressing typological diversity in the context of empirical NLP work.}}
While we focus on achieving high typological distance, our framework can be used with \emph{any} type of information that one wishes to make `diverse' generalizable claims about.
Furthermore, our method can also be used for finding typologically \emph{similar} languages, e.g., with use cases in transfer learning scenarios. 
We recommend that future work in NLP aiming for typological generalizability use our informed selection method (while not losing sight of the human-centered perspective).

\newpage

\appendix

\appendixsection{Sampling evaluation by $k$ for intersection of sampling frames.}\label{app:intersection}

\begin{figure}[H]
    \includegraphics[width=\textwidth]{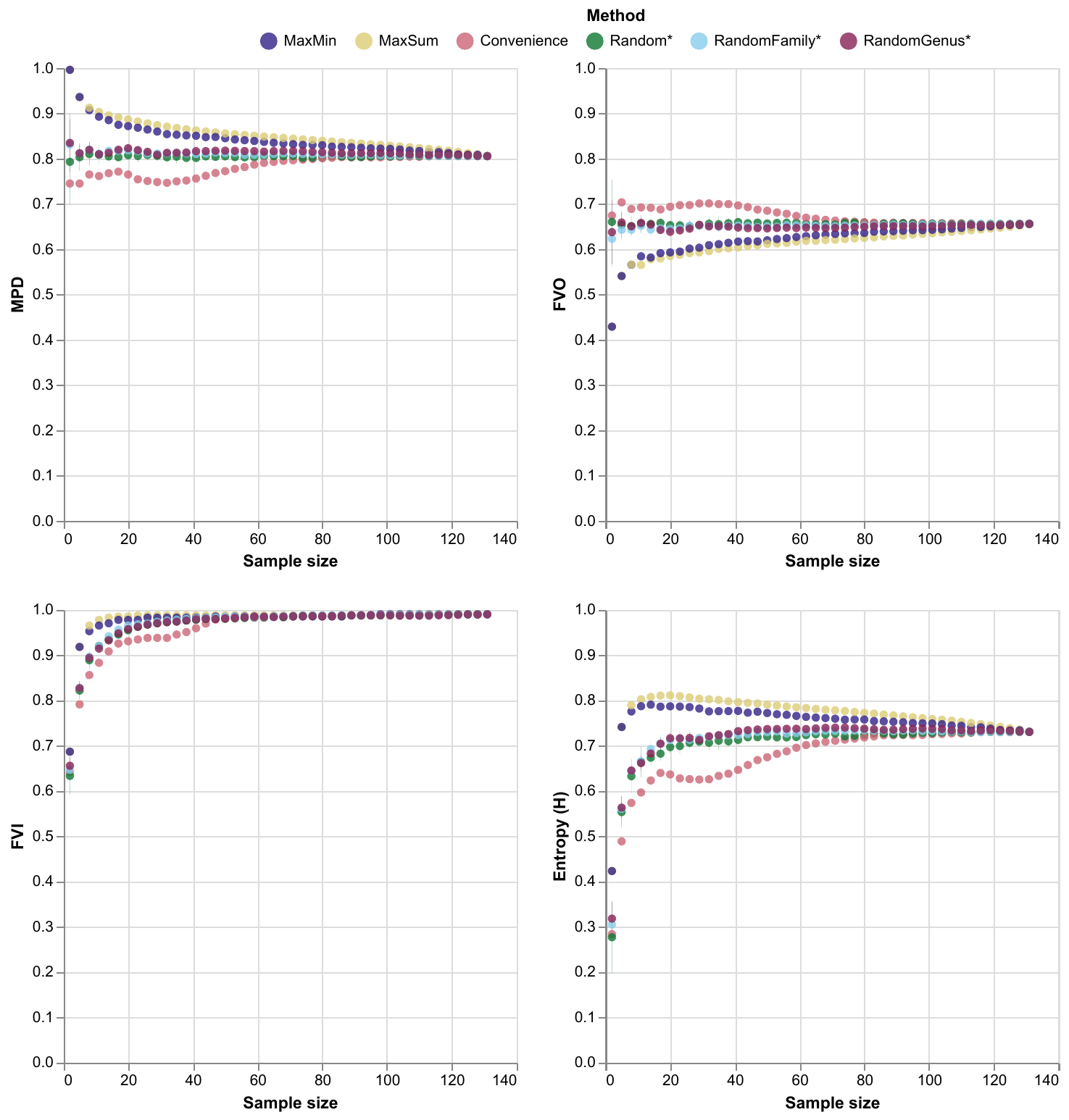}
    \caption{MPD ($\uparrow$), FVO ($\downarrow$), FVI ($\uparrow$) and $H$ ($\uparrow$) for different sample sizes where the sampling frame is equal (intersection for all methods).
    Non-deterministic methods are indicated with an asterisk and averaged over 10 runs, error bars represent their standard deviation. Colors are from Paul Tol's color-blind safe \textit{muted qualitative} scheme.}
    \label{fig:metrics-by-k-intersection-frame}
\end{figure}

\clearpage
\appendixsection{Diversity of downstream sampling frames.}\label{app:frame-div}

\begin{figure}[h]
    \centering
\includegraphics[height=22em]{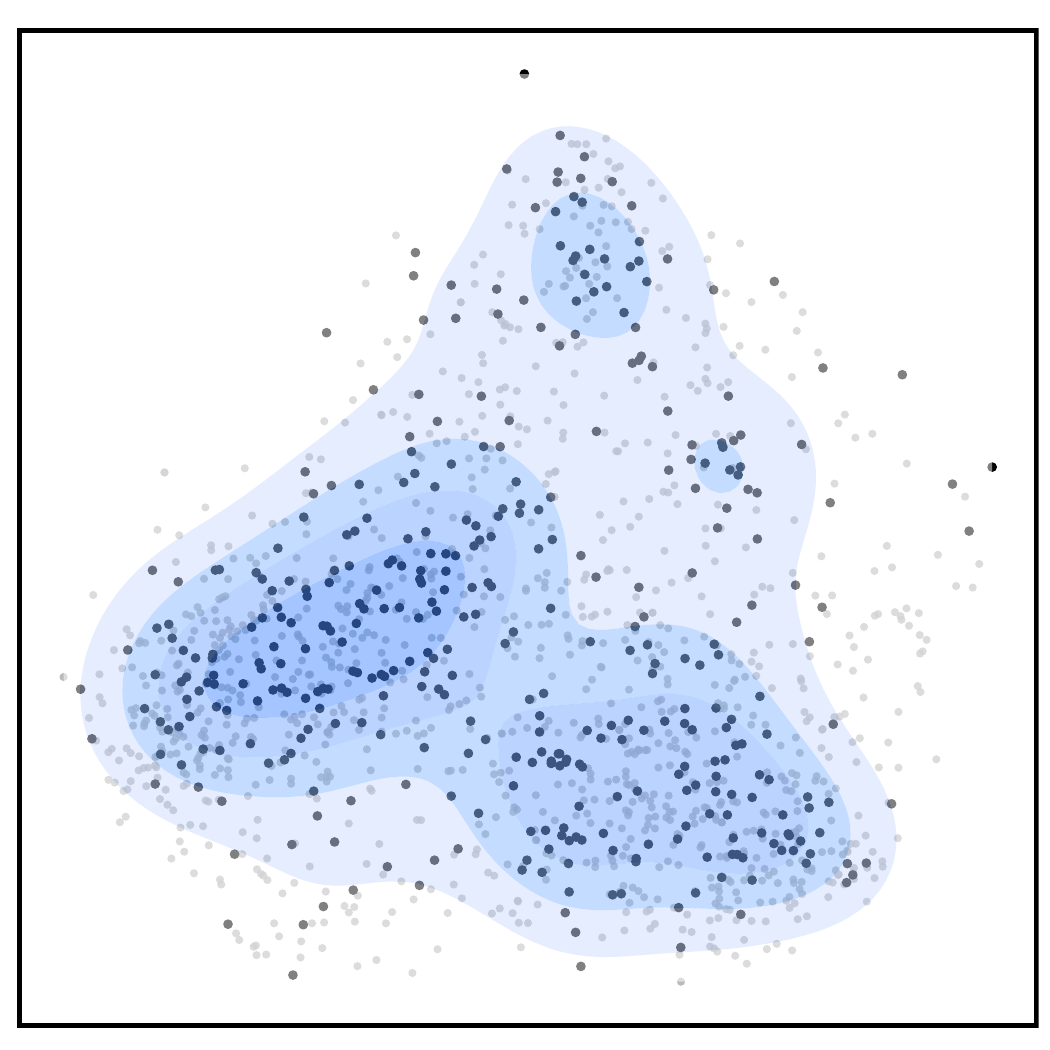}
\caption{\textcolor{major-change}{PCA plot illustrating the extent to which Taxi1500 covers Grambank's `typological space'.}}
\vspace{1em}
\includegraphics[height=22em]{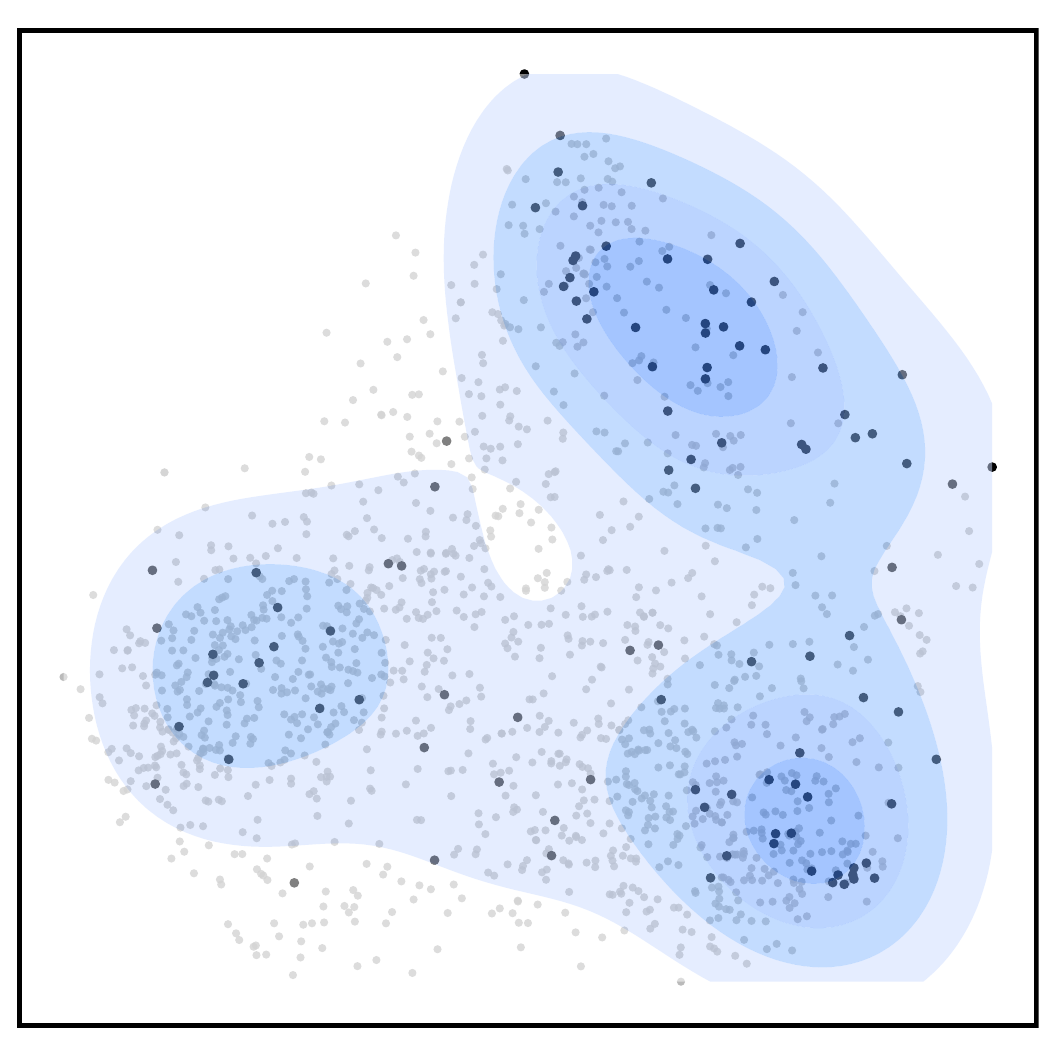}
\caption{\textcolor{major-change}{PCA plot illustrating the extent to which FloRes-200 covers Grambank's `typological space'.}}\label{fig:pca-div}
\end{figure}

\clearpage
\appendixsection{Downstream results on pre-training languages only.}\label{app:train-lang-res}

 \begin{figure}[h]
    \centering
\includegraphics[height=15em]
{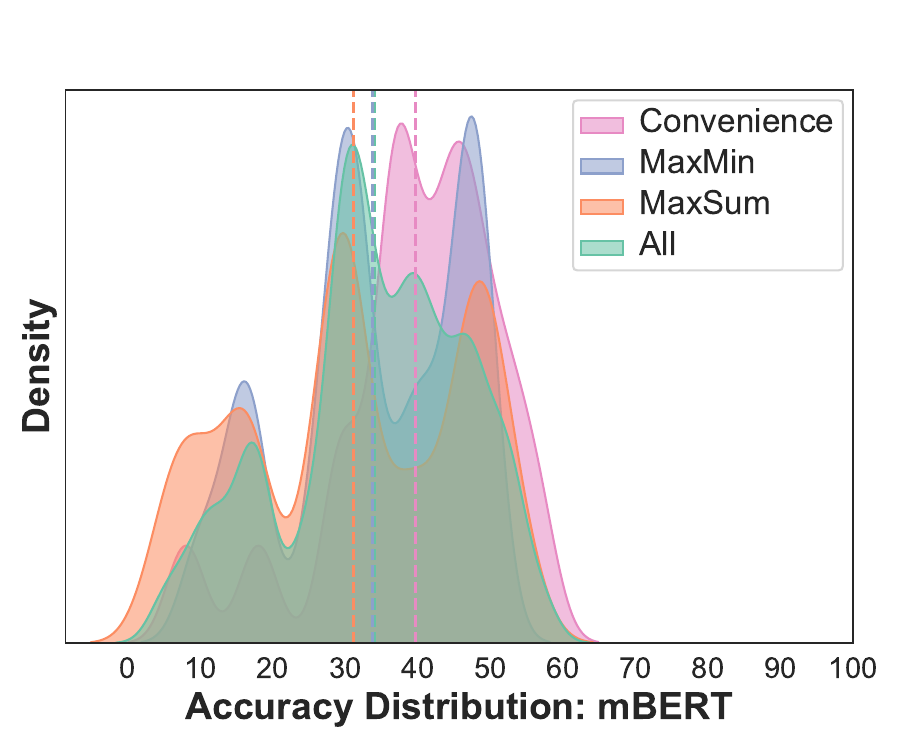}
    \hspace{1em}
\includegraphics[height=15em]{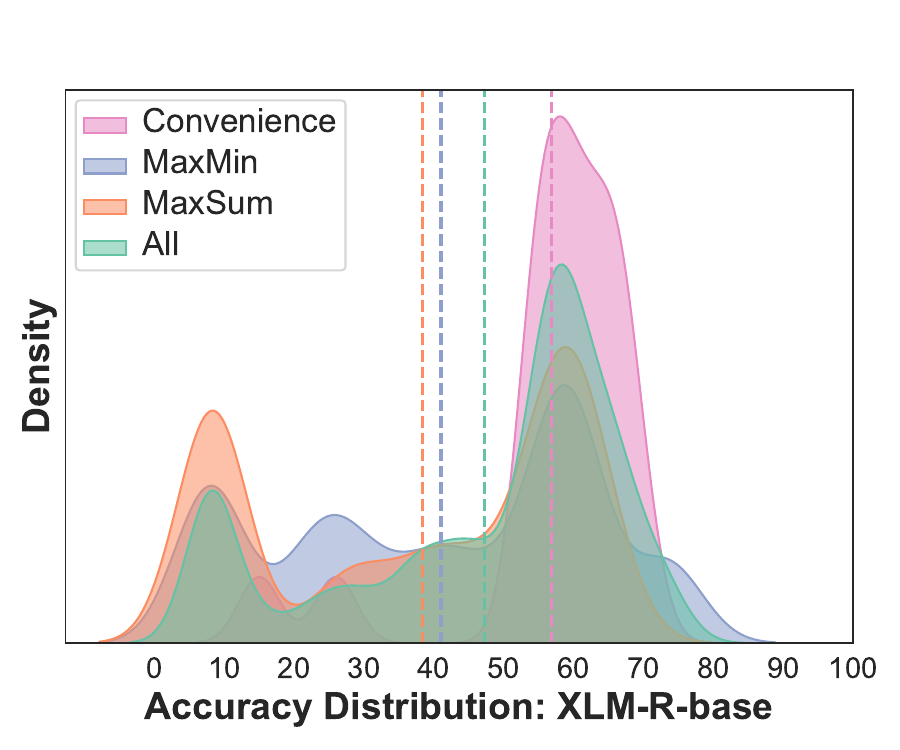}
    \hspace{1em}
\includegraphics[height=15em]{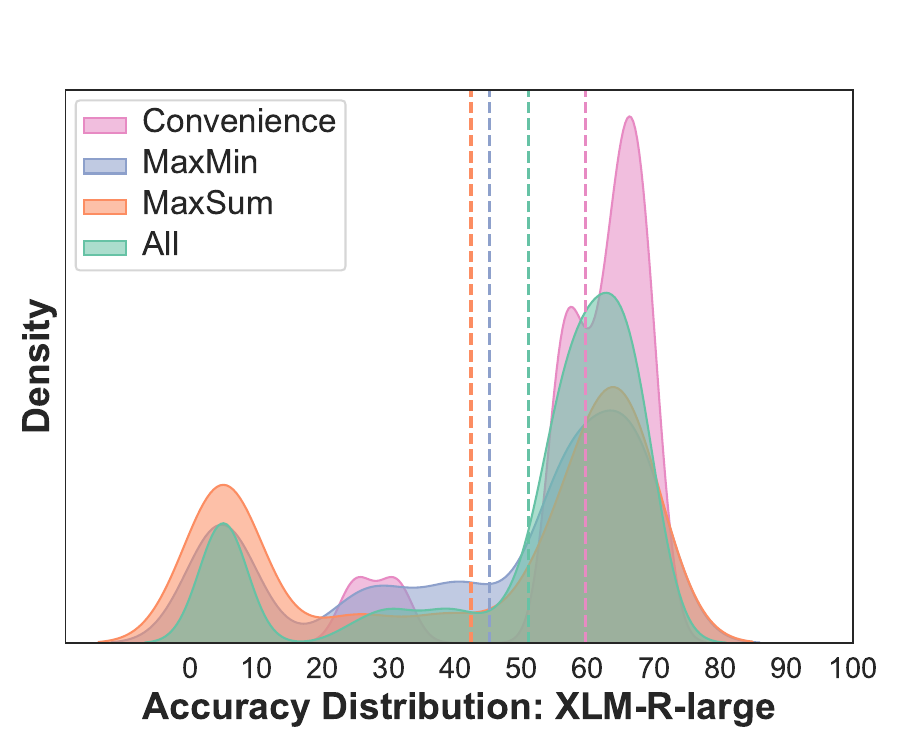}
    \hspace{1em}
\includegraphics[height=15em]{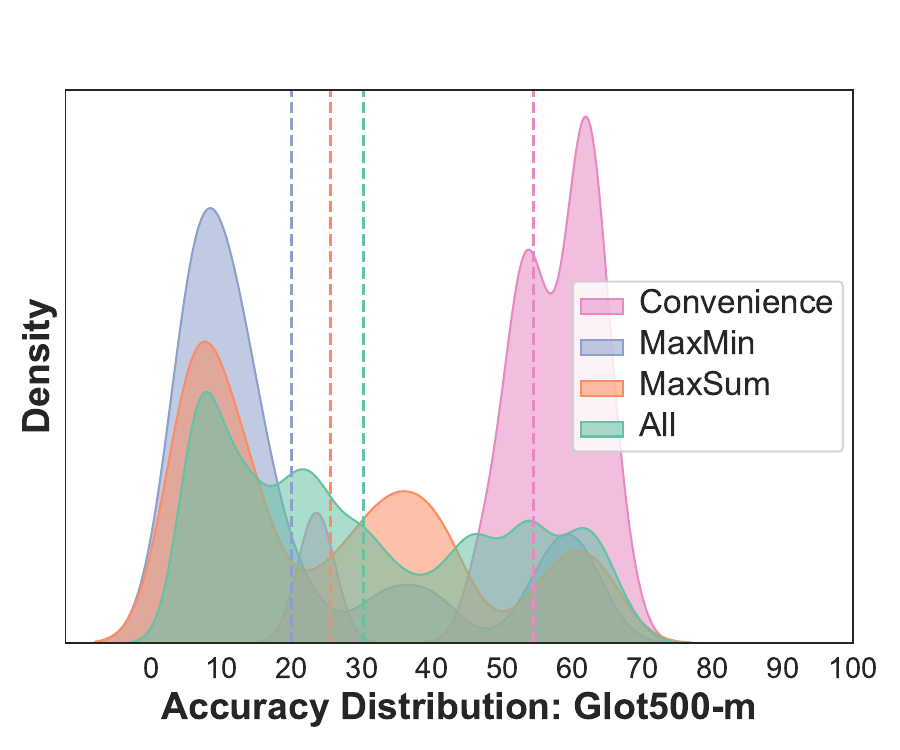}
\caption{\textcolor{major-change-v3}{Accuracy distributions zero-shot performance of language samples ($k=20$) on the Taxi1500 evaluation set, contrasting all four deterministic sampling methods, when sampling only from the pre-training languages of each model.}}\label{fig:taxi1500-results}
\end{figure}

\newpage
\starttwocolumn

\begin{acknowledgments}
We thank the members of LAGoM-NLP at KU Leuven, the AAU-NLP lab at Aalborg University, and the Helsinki-NLP group at the University of Helsinki for feedback on earlier versions of this paper.
We thank Robert Östling for pointing us to Anna Sjöberg's thesis.
We thank Kaius Sinnemäki and Anna Sjöberg for discussing the scope of this project with us from a typological perspective.
We thank Hedvig Skirgård for helpful feedback on cropping the dataset according to coverage, removing macrolanguages, and the overall scope of this paper.
Any remaining errors are our own.
EP and JB are funded by the Carlsberg Foundation, under the \textit{Semper Ardens: Accelerate} programme (project nr. CF21-0454).
WP is funded by a KU Leuven Bijzonder Onderzoeksfonds C1 project with reference C14/23/096.
This work received support from the CA21167 COST action UniDive, funded by COST (European Cooperation in Science and Technology).
\end{acknowledgments}

\bibliography{main}

\end{document}